\pdfoutput=1
\documentclass[11pt]{article}

\usepackage[final]{acl}

\usepackage{times}
\usepackage{mathtools}
\usepackage{placeins}
\usepackage{latexsym}
\usepackage{amssymb}
\usepackage{tcolorbox}
\usepackage{parcolumns}
\usepackage{subcaption}
\usepackage{multirow}
\usepackage[T1]{fontenc}
\usepackage[utf8]{inputenc}

\usepackage{microtype}

\usepackage{inconsolata}
\usepackage{amsmath}
\usepackage{longtable}
\usepackage{dblfloatfix}
\usepackage{placeins}  

\usepackage{xspace}
\usepackage{graphicx}
\usepackage{booktabs}

\newcommand{\ourmodel}{\textsc{Score}\xspace}

\title{\ourmodel: Specificity, Context Utilization, Robustness, and Relevance for Reference-Free LLM Evaluation}
\author{
  Homaira Huda Shomee$^{1,2}$ \quad
  Rochana Chaturvedi$^{2}$ \quad
  Yangxinyu Xie$^{2,3}$ \quad
  Tanwi Mallick$^{2}$ \\
  $^1$University of Illinois Chicago, Chicago, IL, USA \\
  $^2$Argonne National Laboratory, Lemont, IL, USA \\
  $^3$University of Pennsylvania, Philadelphia, PA, USA \\
  \texttt{\{hshome2@uic.edu, rchaturvedi@anl.gov, xinyux@wharton.upenn.edu, tmallick@anl.gov\}}
}

\begin{document}
\maketitle
\begin{abstract}
Large language models (LLMs) are increasingly used to support question answering and decision-making in high-stakes, domain-specific settings such as natural hazard response and infrastructure planning, where effective answers must convey fine-grained, decision-critical details. However, existing evaluation frameworks for retrieval-augmented generation (RAG) and open-ended question answering primarily rely on surface-level similarity, factual consistency, or semantic relevance, and often fail to assess whether responses provide the specific information required for domain-sensitive decisions. To address this gap, we propose a multi-dimensional, reference-free evaluation framework that  assesses LLM outputs along four complementary dimensions: specificity, robustness to paraphrasing and semantic perturbations, answer relevance, and context utilization. We introduce a curated dataset of 1,412 domain-specific question–answer pairs spanning 40 professional roles and seven natural hazard types to support systematic evaluation. We further conduct human evaluation to assess inter-annotator agreement and alignment between model outputs and human judgments, which highlights the inherent subjectivity of open-ended, domain-specific evaluation. Our results show that no single metric sufficiently captures answer quality in isolation and demonstrate the need for structured, multi-metric evaluation frameworks when deploying LLMs in high-stakes applications. 
% Large language models (LLMs) are increasingly used in domain-specific, high-stakes question answering, where accurate and complete representation of fine-grained details is critical. 
% However, existing evaluation frameworks for retrieval-augmented generation (RAG) and open-ended QA often fail to assess whether responses contain precise and specific information. We propose a multi-dimensional, reference-free evaluation framework that  assesses LLM outputs along four complementary dimensions: specificity, robustness to paraphrasing and semantic perturbations, answer relevance, and context utilization. To support systematic evaluation, we introduce a curated dataset of 1,412 domain-specific question-answer pairs spanning 40 professions and seven natural hazard types. Using this dataset, we evaluate multiple state-of-the-art LLMs under different evaluator settings. We further conduct human evaluation to assess inter-annotator agreement and alignment between model outputs and human judgments, which highlights the inherent subjectivity of open-ended, domain-specific evaluation. Our results show that no single metric sufficiently captures answer quality in isolation and demonstrate the need for structured, multi-metric evaluation frameworks when deploying LLMs in high-stakes applications.
% \todo{Write abstarct};;;
% \sm{the title should say something about domain or dataset; you have two main contributions: dataset and evaluation framework}

\end{abstract}

\section{Introduction}
% \sm{Not all my comments are in red; there are some as overleaf comments on the side bar}
LLMs have shown remarkable capabilities across a wide range of tasks, specially in answer generation and decision support \cite{brown2020language, ouyang2022training,wei2022emergent}. As these models are increasingly deployed in high-stakes domains such as healthcare, legal reasoning, and climate risk assessment \cite{wu2025medical,nagar2024umedsum,li2024legalagentbench, luo2025automating,li2024cllmate}; the need for robust and reliable evaluation methods has become essential. One such critical domain is natural hazard response, where LLMs are used to recommend mitigation strategies and provide factual information about disasters in specific geographic regions \cite{xie2025marsha}. In such scenarios, even minor inaccuracies can lead to real-world consequences.

Despite this growing reliance on LLMs, evaluating their outputs in the above mentioned settings remains challenging. Traditional reference-based evaluation metrics are often insufficient, as they primarily measure surface-level similarity to gold-standard answer rather than verifying whether generated responses contain correct and specific information \cite{celikyilmaz2020evaluation,kryscinski2020evaluating,chang2024survey}. This limitation is particularly acute in  natural hazard and extreme weather response, where publicly available gold-standard datasets are scarce due to the diversity of hazards, geographic regions, infrastructure systems, and professional contexts. Effective answers in this domain must be geographically precise, and tailored to specific concerns. 
% These challenges highlight the need for both a domain-specific evaluation dataset and reference-free evaluation methods capable of assessing answer quality along multiple dimensions.

Recent work has shifted toward reference-free evaluation, including LLM-as-a-judge pipelines, where LLMs themselves act as evaluators. Early frameworks like MT-Bench \cite{bai2024mt} and AlpacaEval \cite{alpacaeval2023} pioneered the use of high-capacity models like GPT-4 to score multi-turn dialogues or perform pairwise comparisons, with later variants adjusting for verbosity bias. Chatbot Arena \cite{chiang2024chatbot} leverages large-scale human preferences to rank models via Elo scores. While these benchmarks primarily focus on general-purpose conversational ability, they are not designed to capture the level of fine-grained specificity and answer relevance required in high-stakes applications. In parallel, the rise of Retrieval-Augmented Generation (RAG) has motivated evaluation frameworks that focus on factuality, answer relevance, claim-level verification, and context relevance or utilization \citep{es2024ragas, ru2024ragchecker, saad-falcon-etal-2024-ares, ni2024afacta}.

% As LLM-based systems increasingly incorporate external knowledge sources, research focus has shifted toward evaluating grounding and factuality within Retrieval-Augmented Generation (RAG) workflows. Existing RAG evaluation frameworks can be broadly grouped into four categories. First, faithfulness or factuality metrics to assess whether generated answers are supported by retrieved documents \cite{es2024ragas, ru2024ragchecker}. Second, answer relevance metrics evaluate whether responses directly address the user’s query \cite{es2024ragas,saad-falcon-etal-2024-ares}. Third, claim-level verification approaches decompose answers into atomic claims and assess their factual validity against evidence \cite{ni2024afacta}.  Fourth, context relevance or utilization metrics aim to measure the extent to which retrieved knowledge is used to generate the answer \cite{es2024ragas}.

Although effective for general settings, existing RAG evaluation metrics fall short in domain-specific settings. Faithfulness and factuality metrics assess whether answers are broadly supported by retrieved documents, but does not verify whether critical specific details are present and correct \cite{min2023factscore}. Answer relevance metrics mostly rely on semantic similarity or LLM-based scores, which can overestimate quality in cases where responses are generic rather than informative and relevant \cite{liu2023g}. Claim-level verification focuses on factual correctness at the level of individual claims, but do not evaluate whether the answer sufficiently addresses the user’s query intent or decision context. Finally, for context utilization, prior work \citep{ru2024ragchecker} measures the claim-level semantic overlap between retrieved knowledge and generated answer. While this captures whether the generated answer is supported by the retrieved information, it does not measure whether the model actually relied on that information during generation as LLMs may produce overlapping claims from their parametric knowledge.

To address these limitations, we propose \ourmodel, a multi-perspective evaluation framework for assessing LLM-generated answers in natural hazard analysis and decision support. Our framework evaluates responses along five dimensions: \textit{specificity}, \textit{relevance}, \textit{robustness}, \textit{context utilization}, and \textit{readability}. For specificity, we employ multiple LLM-as-a-judge setup to verify whether key details (e.g., hazard type, location, timeline, intensity) are explicitly stated and supported by retrieved evidence, and aggregate the judges’ decisions into a final score. We assess robustness by applying paraphrasing and controlled perturbations to the question. By measuring the consistency of the model’s responses across these variations, we detect over-sensitivity in model behavior. For answer relevance, we adopt a question-regeneration strategy with an additional semantic masking step that removes domain-specific entities. Given the use of retrieval-augmented generation, we evaluate context utilization by measuring whether and how retrieved passages contribute to the generated answer. Our metric measures evidence utilization at a finer granularity through counterfactual dependence by removing individual claims and quantifying the resulting change in answer confidence, enabling us to distinguish necessary evidence from merely consistent, ignored, or distracting context. Finally, we incorporate readability metric using established grade-level metrics (Appendix~\ref{app:readability}) to ensure that responses are appropriate for professional users.

In this work, we present a reference-free evaluation framework \ourmodel for assessing LLMs outputs in high-stakes hazard response. Our key contributions are as follows: (1) We construct a synthetic dataset of 1412 domain-specific question–answer pairs that cover different hazards, locations, infrastructure types, and concerns to support controlled and scalable evaluation; (2) We design a multi-dimensional, reference-free evaluation framework that assesses specificity, robustness, answer relevance, and context utilization, and validate these metrics through human evaluation.%\footnote{Our code and curated data are available at \url{https://anonymous.4open.science/r/score-8D5F/readme.md}} 

\textbf{Code \& Data:} Our code and curated data are available at \url{https://anonymous.4open.science/r/score-8D5F/readme.md}
% \clearpage
% \input{latex/2_Background}
% \clearpage
% \input{latex/2_data.tex}
\section{Dataset Construction}
\label{sec:dataset}
% \todo{Shomee: can we replace the word "revised: did prior work have this, can we say a hazard-location mapping grounded in real-world context}
We construct the largest dataset to date of infrastructure-related questions that simulate real-world hazard scenarios and decision-making contexts. Recent work by \citet{xie2025marsha} demonstrates the effectiveness of agentic frameworks for addressing critical infrastructure risks by incorporating structured user profiles that encode professional background and user concerns. While their work focuses on a conversational retrieval-augmented generation (RAG) system for wildfire risk mitigation, we extend similar role-aware design principles to develop a comprehensive evaluation benchmark that systematically assesses whether LLMs generate contextually appropriate responses across a broader range of infrastructure sectors, hazard types, and geographic settings.%Recent work by \citet{xie2025marsha} demonstrated the effectiveness of agentic frameworks for addressing critical infrastructure risks by incorporating structured user profiles that encode professional background and user concern. While \citet{xie2025marsha} focused on building a conversational retrieval-augmented generation (RAG) architecture for wildfire risk mitigation, we extend the role-aware design principles into constructing a comprehensive evaluation benchmark that systematically assesses whether LLMs can generate contextually appropriate responses across more diverse infrastructure sectors, hazard types, and geographic settings.
\paragraph{User Profile. } Each question-answer is associated with a synthetic user profile comprising \textit{profession}, \textit{concern} (fact-based, recommendation-seeking, or hybrid), \textit{location}, and \textit{timeline}. Profession and concern characterize a user's role and objective, while location and timeline capture the geographic scope and time frame. For each question, profile attributes are instantiated through controlled randomization: first a hazard type is sampled, then a location is sampled conditional on the hazard, then the remaining attributes (profession, and timeline) are randomly selected from predefined sets.
%The dataset is generated using a structured, template-based pipeline that incorporates both professional and hazard-specific context, inspired by \citet{xie2025marsha}, and introduces a more comprehensive set of hazard types and revised hazard–location mappings.\todo{We are making changes here}%, and large-scale dataset construction.\\
% \sm{is it really large scale? you can say the largest data in this domain}
\paragraph{Professional Context.} To ground questions in real-world professional contexts, we curate a diverse set of 40 professions spanning five critical infrastructure sectors: \textit{transportation}, \textit{water}, \textit{energy}, \textit{buildings}, and \textit{communication} (Appendix Table \ref{tab:profession}). Each corresponds to roles that are actively involved in hazard mitigation and infrastructure planning, such as, transportation planner, hydraulic engineer, energy storage specialist, building systems manager, and telecommunications engineer.
\paragraph{Location and Hazard Type.}
We select the location and hazard types based on realistic hazard-region mappings that frequently occur within the U.S. context. We identify a set of seven hazard types: \textit{cold wave}, \textit{heat wave}, \textit{coastal flooding}, \textit{ice storm}, \textit{hurricane}, \textit{drought}, and \textit{wildfire}. Each hazard is then associated with a curated list of U.S. locations that are historically or geographically susceptible to it, based on Natural Risk Index(Appendix Table \ref{tab:hazard_location}).\footnote{https://hazards.fema.gov/nri/data-resources} For instance, hurricanes are commonly associated with Miami or New Orleans, while wildfire scenarios with San Diego or Elko, Nevada. This helps ensure that the generated questions reflect regionally relevant risks and real-world challenges faced by infrastructure planners.% Additional details are provided in Table \ref{tab:hazard_location} in App. \ref{app:data}.\\
\paragraph{Question Generation. } Each template has placeholders for infrastructure, hazard, location, concern, and timeline, populated using user profile attributes. To create a diverse set of questions, we include three concerns: fact-based, recommendation-seeking, and hybrid. Fact-based templates focus on objective inquiries (e.g., ``What are the critical vulnerabilities of [INFRASTRUCTURE] to [HAZARD] in [LOCATION]?''). In contrast, recommendation-seeking templates are designed to elicit forward-looking or strategic guidance (e.g., ``What strategies should [PROFESSION] consider to enhance [INFRASTRUCTURE] resilience in the face of [HAZARD] over the next [TIMELINE] years?''). Hybrid templates focus on infrastructure interdependencies and cascading impacts across sectors (e.g., ``What are the cascading impacts of [HAZARD] between [INFRASTRUCTURE] and building systems in [LOCATION]?''). During generation, the concern type in the user profile determines the template pool, from which one template is randomly selected.
\paragraph{Answer Generation.}
For each question, we retrieve the top five semantically relevant abstracts from a domain-specific knowledge base \cite{mallick2023analyzing, mallick2025understanding}. These abstracts, along with the user profile and query, are incorporated into a structured prompt that is passed to an LLM to generate a concise, point-wise answer grounded in the retrieved literature (Appendix \ref{app:ag}).  
% These abstracts serve as the grounding documents for the response. Once the literature is retrieved, we construct a detailed prompt that incorporates the five abstracts, the user’s contextual profile (including profession, concern type, location, timeline, and scope), and the question itself. This prompt is passed to an LLM (e.g., GPT, Gemini) to generate an answer in a concise, point-wise format. The prompt is designed to ensure that the response explicitly utilize the literature and user's context (e.g., profile). 
% \vspace{-15pt}
\section{Evaluation Framework}
A major contribution of our work is the design of a reference-free, multi-dimensional evaluation framework for assessing LLM-generated answers in high-stakes, domain-specific settings. To rigorously evaluate the responses generated in Section \ref{sec:dataset}, our framework measures the answer quality across four fine-grained dimensions: \textbf{specificity}, \textbf{robustness}, \textbf{answer relevance}, and \textbf{context utilization}. Additionally, we asses \textbf{readability} using established metrics to ensure that the outputs are suitable for professional use (Appendix \ref{app:readability}).

\subsection{Specificity}
% \sm{My major comment is that this section (evaluation framework) has a lot of text: be crisp, and make things more technical (like you did add a lot of equations, these might be enough) - details can be referred to appendix. also this gives space for experimental results}
\begin{figure}
    \centering
    \includegraphics[width=0.8\linewidth]{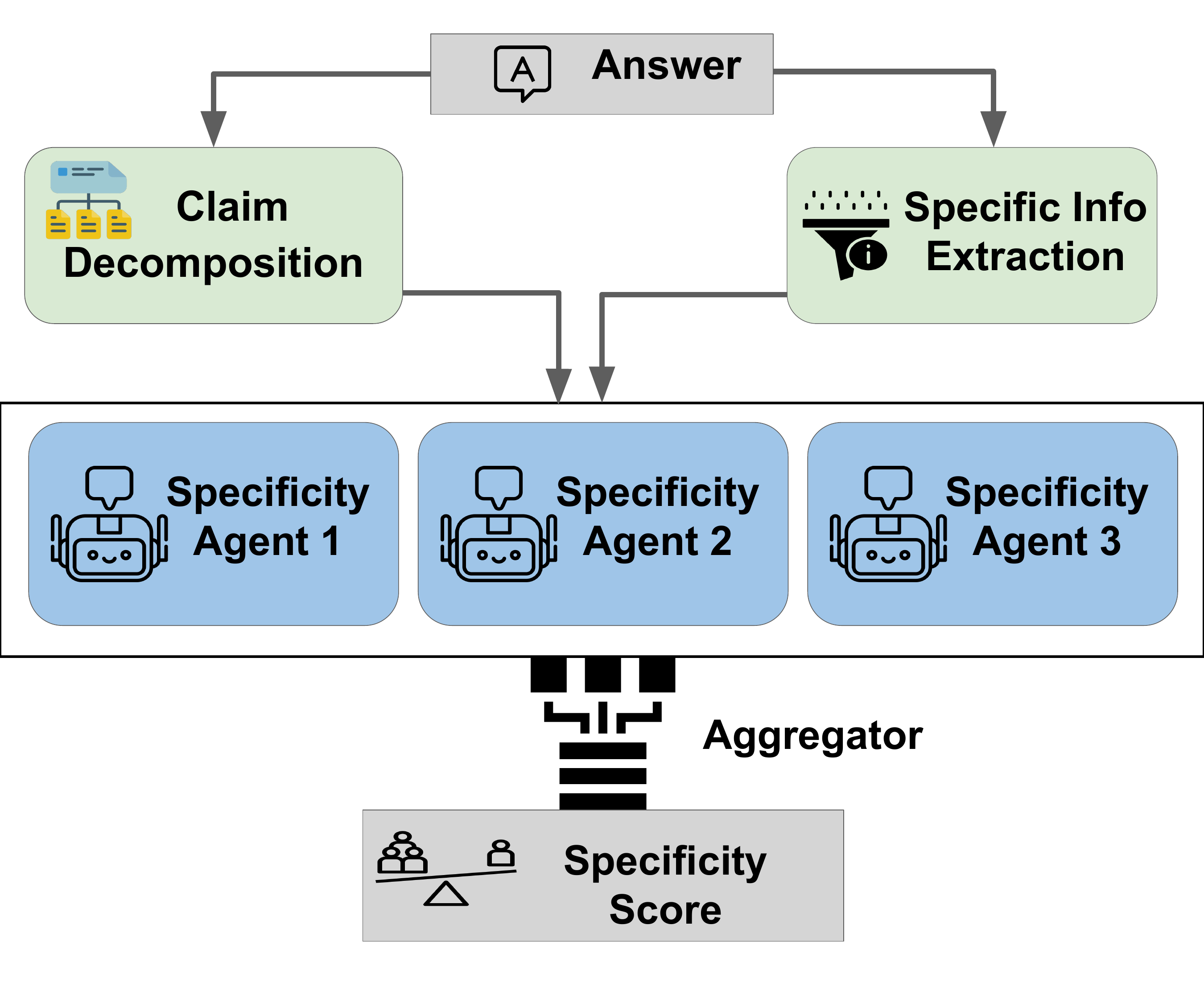}
    \caption{Specificity score computation: each generated answer is decomposed into atomic claims, specific details (hazard type, location, timeline, intensity) are extracted, and each claim is evaluated using multiple LLM-based agents before aggregating their judgments.}
    \label{fig:specificity}
\end{figure}
% We design a multi-agent evaluation framework to assess the specificity of answer generated by LLMs. Our objective is to quantify whether the generated answer provide fine-grained specific details that are contextually relevant to the input query.
We propose a multi-agent evaluation framework, illustrated in Figure \ref{fig:specificity}, to measure the specificity of LLM-generated answers by assessing whether fine-grained details (e.g., hazard type, location, timeline, and intensity) are explicitly stated and verifiable against the retrieved knowledge. Formally, given the set of atomic claims \( C\coloneqq\{c_1, c_2, \ldots, c_|C|\} \) extracted from generated answers, and specificity dimensions \( D\coloneqq\{d_h, d_l, d_t, d_i\} \) representing \textbf{h}azard type, \textbf{l}ocation, \textbf{t}imeline, and \textbf{i}ntensity. An evaluator agent $k$ assigns a label \(\ell_{i d_j}^{(k)} \in \{\text{yes},\ \text{no},\ \text{n/a}\}\) to each claim \(c_i\), along each dimension. Here,
% \[
% A_k(c_i) = \big[\, \ell_{i d_h}^{(k)},\ \ell_{i d_l}^{(k)},\ \ell_{i d_t}^{(k)},\ \ell_{i d_i}^{(k)} \,\big],
% \]
yes indicates that the specific detail is both present in the claim and verifiable using the knowledge source; no indicates a lack of support or contradiction; and n/a denotes that the detail is not mentioned in the claim. We map these categorical labels to numerical values:
\[
s_{i d_j}^{(k)} =
\begin{cases}
1, & \text{if } \ell_{i d_j}^{(k)} = \text{``yes''}, \\[4pt]
0, & \text{if } \ell_{i d_j}^{(k)} = \text{``no''}, \\[4pt]
\text{n/a}, & \text{if } \ell_{i d_j}^{(k)} = \text{``n/a''}.
\end{cases}
\]
For each claim \(c_i\) and
dimension \(d_j\), we do majority voting across $k$ evaluators, to obtain:
\[
\mathbf{s}_i = 
\big[\, {s}_{i d_h},\ 
         {s}_{i d_l},\ 
         {s}_{i d_t},\ 
         {s}_{i d_i} \,\big].
\]
For each dimension \(d_j\), we average the consensus $s_{id_j}$ across claims where  \(d_j\neq \text{n/a}\). This prevents infrequently extracted dimensions (e.g., intensity) from being penalized by averaging over all claims:
\[
\bar{s}_{d_j}
=
\frac{
\sum_{i \in C \,:\, s_{i d_j} \neq \text{n/a}}
s_{i d_j}
}{
\sum_{i \in C \,:\, s_{i d_j} \neq \text{n/a}} 1
}
\]
% This \textit{valid scoring} step ensures that dimensions which appear
% infrequently in the dataset (e.g., intensity) are not unfairly penalized
% by dividing over all claims. \\
% This valid-scoring strategy prevents dimensions that appear infrequently (e.g., intensity) from being penalized by averaging over all claims.
The final specificity score is a weighted combination of dimension-level averages:
\begin{equation}
\text{Specificity}(C)
=
\frac{
\sum_{j:\, \bar{s}_{d_j} }%\text{ defined}}
\alpha_j\, \bar{s}_{d_j}
}{
\sum_{j:\, \bar{s}_{d_j}}% \text{ defined}}
\alpha_j
}
\end{equation}
where,
\(
{\alpha} = [0.6,\ 0.2,\ 0.1,\ 0.1]
\)
corresponds to the ordered dimensions:
\textit{hazard}, \textit{location}, \textit{timeline}, and \textit{intensity}. If a dimension is ``n/a'' across claims, it is excluded from both the numerator and the denominator in the weighted average. In our domain, the four specificity dimensions do not contribute equally to assessing the quality of a claim. We place higher emphasis on the hazard and location dimensions, which are most critical for grounding claims in real-world risk contexts.\footnote{Appendix \ref{app:spec} provides additional details on claim and specific information extraction, and evaluator prompts.}

\subsection{Robustness}
\begin{figure}
    \centering
    \includegraphics[width=0.8\linewidth]{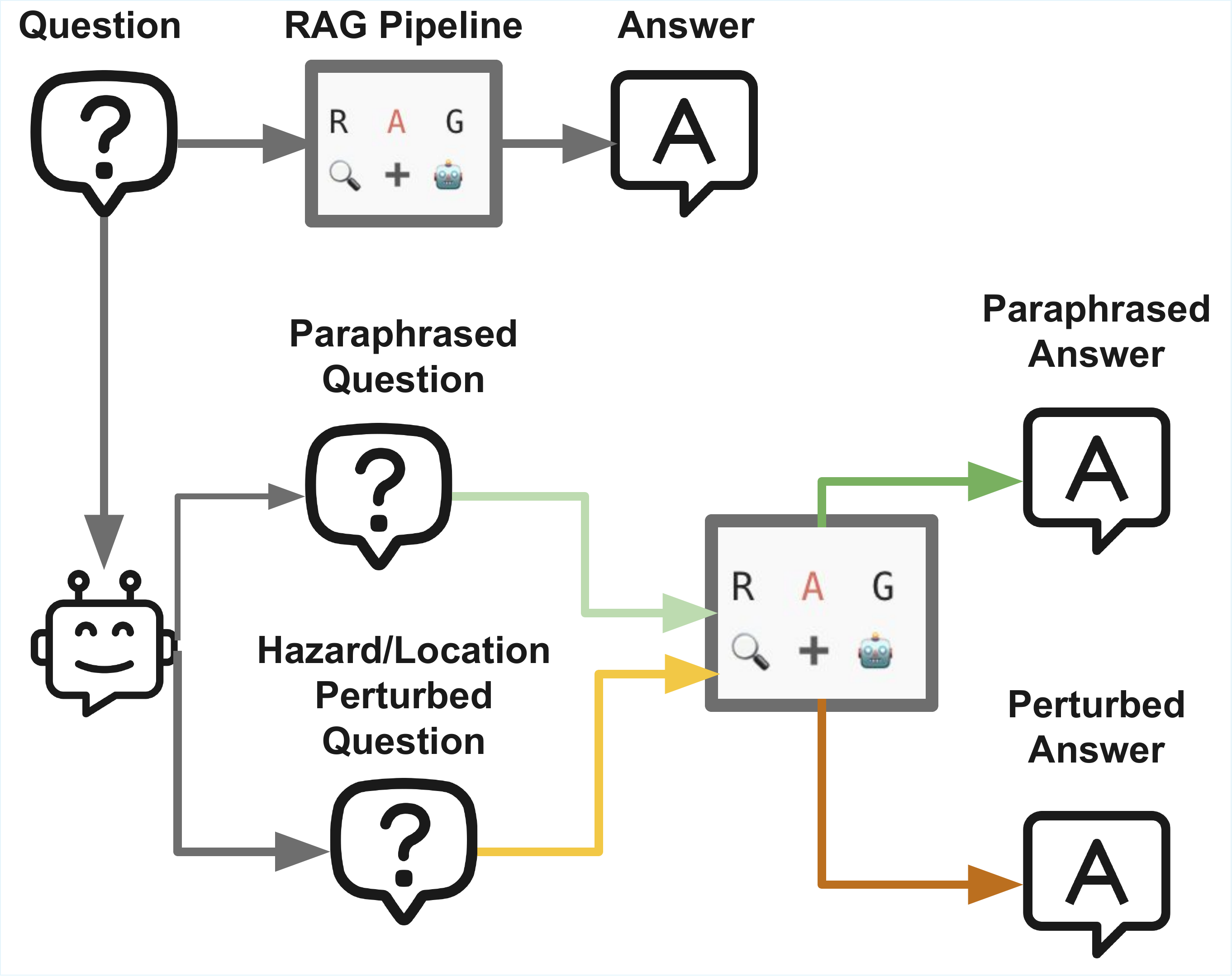}
    \caption{Robustness evaluation workflow. For each question–answer pair, the system generates paraphrased and hazard/location-perturbed variants of the question, runs them through the RAG pipeline, and compares the generated answers to assess semantic consistency and sensitivity to controlled perturbations.}
    \label{fig:robustness}
     \vspace{-10pt}
\end{figure}

This metric assesses the stability of the generated answers by evaluating their consistency under controlled input variations (Figure \ref{fig:robustness}). We examine whether paraphrasing the original question leads to a similar response, and whether intentional changes to key parameters, such as hazard type and/or location, generate appropriately different responses. We consider two types of input modifications:
\paragraph{Paraphrasing:} We rephrase the original question while preserving its semantic meaning. A robust system should generate answers that are semantically consistent with the original response.
\paragraph{Perturbation:} We modify the question by altering the hazard type and/or the geographical location mentioned. In this setting, we expect the system to retrieve a different set of relevant documents and produce correspondingly different answers.
% \sm{the next sentence is obvious/trivial - avoid it}For both scenarios, we regenerate answers using the modified inputs and compare them with the original answer to quantify consistency and sensitivity. 
% Figure  shows the robustness evaluation pipeline.
\subsection{Answer Relevance}
\begin{figure}
    \centering
    \includegraphics[width=0.85\linewidth]{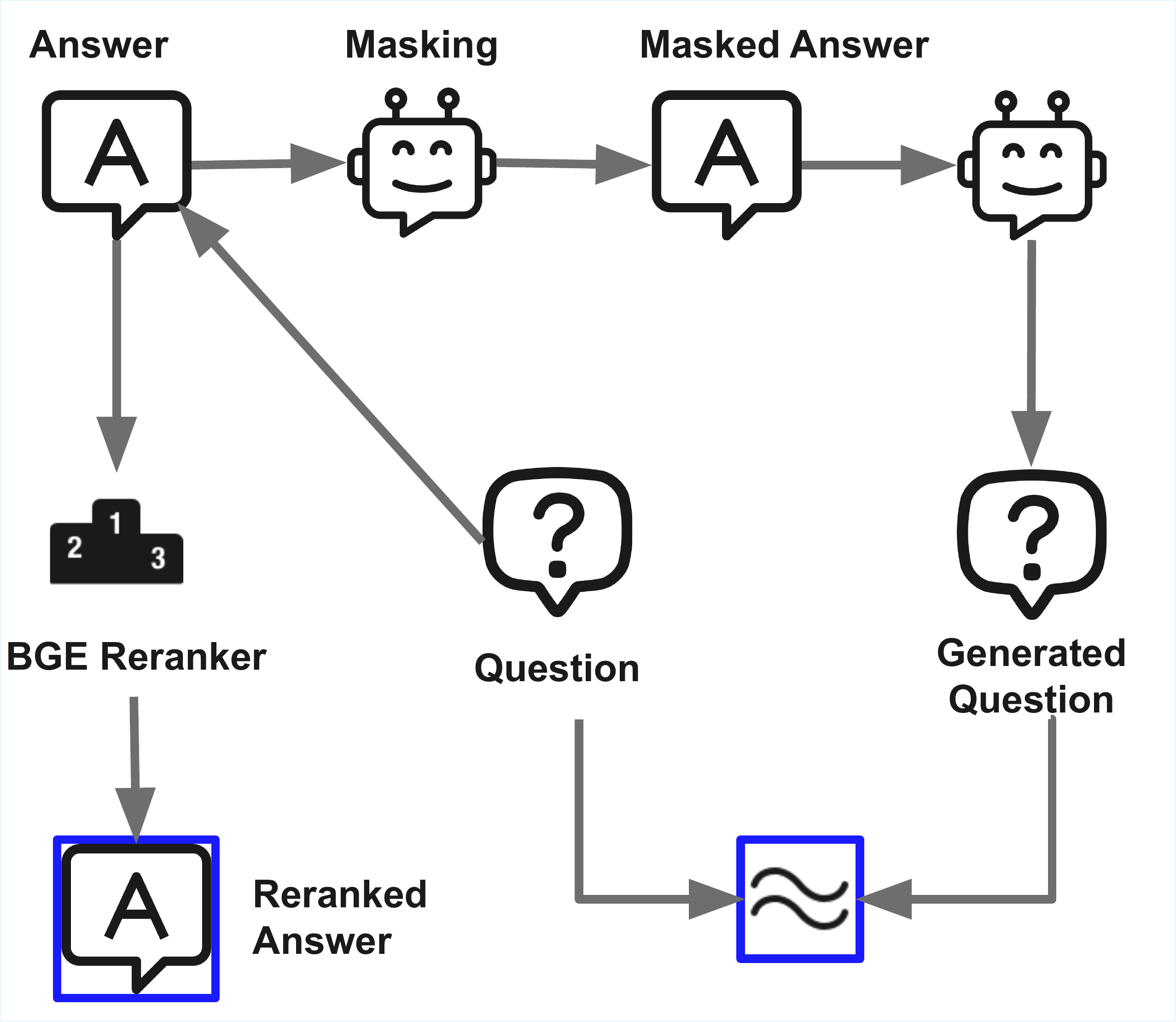}
    \caption{Answer relevance pipeline. For each answer, the system generates reranked answer and answer relevance score. Outputs are shown inside blue boxes.}
    \label{fig:ar}
     \vspace{-10pt}
\end{figure}

Answer relevance evaluates whether a model’s response is relevant to the original user query \cite{saad-falcon-etal-2024-ares, es2024ragas}. We adapt the metric from RAGAS framework \cite{es2024ragas} and extend it by incorporating a masking step to reduce relevance arising from domain-specific lexical overlap. Given a generated answer $a$, we prompt an LLM to generate five candidate questions for which $a$ would be an appropriate answer. This treats LLMs as an inverse question generators. For each generated question, we compute the cosine similarity between its embedding and the embedding of the original question $q$. The final Answer relevance score $R$ is defined as the average similarity. Figure \ref{fig:ar} shows the pipeline.
\paragraph{Answer Relevance with Masking. }High relevance scores can result from superficial lexical overlap, especially when domain-specific keywords (e.g. specific hazard or infrastructure sector) appear in both the question and answer. To mitigate this, we apply a semantic masking step: an LLM first generates a masked answer \( \tilde{a} \) by replacing sensitive tokens with a placeholder (e.g., replacing ``electrical grid” with [INFRASTRUCTURE]), then generates \( \tilde{q} \) from  \( \tilde{a} \) ensuring conceptual rather than lexical alignment. The masked relevance (\( \mathrm{\tilde{R}} \)) score is:
\begin{equation}
\mathrm{\tilde{R}}(q, a) = \frac{1}{n} \sum_{i=1}^{n} \mathrm{sim}\left(q, \tilde{q} \right)
\end{equation}

\noindent\textbf{Relevance Scoring with BGE Reranker. }To complement masked relevance, we apply a post-hoc reranking strategy to assess and enhance the informativeness of multi-point answers. We use a leave-one-out relevance attribution method, where each segment of the answer (e.g., bullet point or paragraph) is evaluated for its marginal contribution to the overall answer quality. We utilize the \texttt{BAAI/bge-reranker-base} model\footnote{\url{https://bge-model.com/tutorial/5_Reranking/5.2.html}}, which provides a dense relevance score \( R(q, a) \) between a question \( Q \) and an answer \( A \). A structured answer is usually composed of an optional introduction and \( m \) content segments $
a\coloneqq\text{Intro} \cup \{a_1, a_2, \dots, a_m\}$.
% The baseline relevance score is computed as:
% \begin{equation*}
% \text{Score}_{\text{full}} = R(q, a)
% \end{equation*}
To estimate the marginal relevance of each individual segment \( a_i \), we remove it from the answer to get $a_{-i}$ and compute the difference:
% \begin{equation*}
%  = a \setminus \{a_i\}
% \end{equation*}
\begin{equation}
\Delta_i =  R(q, a) - R(q, A_{-i})
\end{equation}
% Here, \( \Delta_i \) quantifies the drop in relevance when \( a_i \) is removed. 
A higher \( \Delta_i \) indicates a more significant contribution of that segment to the overall answer quality. If \( \Delta_i < 0 \), it implies that the removed segment was detrimental to the overall relevance.

Finally, we rerank the answer segments \( \{a_i\} \) in descending order of \( \Delta_i \), prepending the introduction (if present). This ensures that the most informative and relevant points appear earlier in the final answer presentation.

\subsection{Context-Utilization}
% \todo{Rochana: push the longer explanations and details to the appendix.}
%"Attribution measures whether or not a chunk affected the response, and Utilization measures how much of the chunk text was involved in the effect."
%usual metrics could be embedding similarity, Token Attribution (which chunks most influenced the output tokens.), LLM-as-a-Judge (might extend hallucinations), human eval (Expensive).
%RAGchecker CU metric: [intersection/entailment between generated claims and relevant retrieved claims]/[intersection/entailment in ground-truth claims and Retrieved relevant information 
% (measured by the number of ground-truth answer claims entailed in retrieved chunks)
We measure context utilization as the degree to which each unique claim in the retrieved context contributes to the model’s final answer. Given a query $q$ and an answer $a$, we first extract a set of unique claims $C\coloneqq\{c_1, c_2, ...c_{|c|}\}$ from the retrieved documents $D$. To quantify the contribution of each claim $c_i\in C$ we adopt a leave-one-out uncertainty–based estimator that measures the change in answer confidence when $c_i$ is removed from the context. Formally, we define the absolute contribution of claim $c_i$ as the drop in model's confidence in generating the answer $a$:
\begin{equation*}
    % A_i = 1-g(a|q\cup C, a|q\cup C\setminus\{c_i\})\\
    \Delta_i(a, q, C, \theta) =  f_\theta(a|q\cup C) - f_\theta(a|q\cup C\setminus\{c_i\})
\end{equation*}
where $f_\theta(a|.)$ denotes the confidence score of language model in generating $a$ conditioned on  $q,C$. We further define relative context utilization as:
\begin{equation*}
    \delta_i(a, q, C, \theta) = \Delta_i(a, q, C, \theta)/{ f_\theta(a|q \cup C)}
\end{equation*}

\noindent\textbf{Confidence Estimation. }We compute answer confidence using the normalized per-token probability of the generated sequence \citep{murray2018correcting}. Specifically, confidence is defined as the geometric mean of token probabilities:% (exponentiation over token-level log probabilities of the sequence normalized by the length of the sequence \citep{murray2018correcting}). %use log prob to avoid multiplication with zero
% \begin{align*}
\begin{equation}
 f_\theta(a|X)=e^{\frac{1}{T}\sum_{t=1}^{T}{log(P(y_t|X, y_{<t}}))}
% \end{align*}
\end{equation}

Where $a\coloneqq(y_1, y_2, ...y_T)$ denotes the answer token sequence, and $X$ denote the conditioning context ($q\cup C$). This formulation avoids length bias and numerical instability associated with raw probability multiplication.\footnote{We also consider token-level predictive entropy.} Using this, $\Delta_i\in(-1,1)$, with larger positive values indicating greater reliance on claim $c_i$. $\delta_i \in (-\infty,1)$, where positive values indicate that the claim contributes positively to the answer, while negative values suggest that the claim is distracting or detrimental.

Finally, we aggregate claim-level contributions to obtain a global context utilization score that summarizes how effectively the retrieved context is used by the model when producing the answer. It is defined as the average ($CU$) or relative ($CU_{rel}$) contribution across all claims:
\begin{equation*}
    CU =\frac{1}{|C|}\sum_{i\in|C|}{\Delta _i}; \quad
    CU_{rel} =\frac{1}{|C|}\sum_{i\in|C|}{\delta _i}
\end{equation*}
% \todo{Shomee: please see this paragraph: if we should move it to intro/motivation}
% Prior work \citep{ru2024ragchecker} measures context utilization as the claim-level semantic overlap between retrieved knowledge and generated answer \citep{ru2024ragchecker}. While this captures whether the generated answer is supported by the retrieved information, it does not measure whether the model actually relied on that information during generation as LLMs may produce answers from their parametric knowledge making claims overlapping with this knowledge unused. In contrast, our metric measures evidence utilization at a finer granularity through counterfactual dependence by removing individual claims and quantifying the resulting change in answer confidence, enabling us to distinguish necessary evidence from merely consistent or ignored or distracting context.
  
 % todo3: check extreme examples: around 0, 1, -1

% \subsection{Profile consideration}
% \clearpage
\begin{table*}[ht]
\centering
\resizebox{0.8\textwidth}{!}{%
\begin{tabular}{cccccc}
\toprule
\textbf{Generator}& \textbf{Evaluator} & \textbf{Specificity} $\uparrow$ & \multicolumn{2}{c}{\textbf{Robustness} } &    \textbf{Answer Relevance} $\uparrow$ \\
\cmidrule(lr){4-5}
& & & \textbf{Para.}$\uparrow$ & \textbf{Pert.}$\downarrow$ &  \\
\midrule
\multirow{2}{*}{GPT-4o }&GPT-4o &   $0.61 \pm 0.21 $   &  $ 0.89\pm0.05  $ &$ 0.60  \pm  0.09$   &$   0.69 \pm  0.11 $    \\

 & Qwen3-8B  &  $0.58 \pm 0.22 $     &  $ 0.89\pm 0.04 $ & $ 0.66\pm 0.10 $    & $ 0.73\pm0.18  $          \\

\midrule
\multirow{2}{*}{Gemini2.5} & GPT-4o & $0.55 \pm 0.24 $    & $ 0.87\pm 0.07 $   &  $ 0.87\pm 0.07$      &  $ 0.68\pm 0.1 $     \\
 & Qwen3-8B &   $0.45 \pm  0.27$     &  $0.86 \pm  0.05$   &  $0.66 \pm0.10  $   & $ 0.83\pm 0.11  $  \\  

\bottomrule
\end{tabular}}
\caption{Evaluation results for two different generators (GPT-4o and Gemini2.5) with mean and standard deviation of all measures for each generator-evaluator pair. Higher values indicate better performance for all metrics except perturbation robustness, where lower values show higher sensitivity to changes. Overall, GPT-4o produces more detailed and specific responses than Gemini 2.5. The evaluation results indicate that GPT-4o acts as a more lenient judge than Qwen3-8B.}
 \vspace{-10pt}
%Overall,  GPT-4o produces more detailed and specific responses than Gemini 2.5. Gemini achieves higher answer relevance scores, reflecting a trade-off between detailed and more generic responses. The evaluation results also indicate that GPT-4o acts as a more lenient judge, whereas Qwen3-8B consistently assigns lower and stricter specificity scores across both generators.}
% Overall, the results show that GPT-4o produces more detailed and specific responses than Gemini 2.5. Gemini achieves higher answer relevance scores, reflecting a trade-off between detailed and more generic responses. The evaluation results also indicate that GPT-4o acts as a more lenient judge, whereas Qwen3-8B consistently assigns lower and stricter specificity scores across both generators.
% \sm{always add a conclusion sentence}}
\label{tab:specificity}
\end{table*}

\section{Experimental Setup}
In our experimental setup, LLMs are used as (i) answer generators within a RAG pipeline, and (ii) evaluators for the proposed metrics. We employ GPT-4o and Gemini 2.5 as the primary answer generators. These models use the RAG pipeline described, where the top five semantically relevant abstracts from our domain-specific knowledge base serve as grounding per query. For the evaluation phase, we use both proprietary and open-source models to show the framework's flexibility across resource tiers. Specifically, we employ GPT-4o as a commercial evaluator and Qwen3-8B as a representative open-source evaluator. We assess answer relevance by incorporating the BAAI/bge-reranker-base model to compute relevance scores for multi-point answers. Proprietary models like GPT-4o and Gemini do not expose token-level logits, needed to estimate CU score. We decouple answer generation and confidence estimation using teacher-forcing on lightweight models. We use LLaMA3.1-8B-Instruct, Qwen2.5-7B-Instruct, and Qwen3-8B as proxies. We also experiment with Gemma2-9B-it and Ministral-8B-Instruct. See Appendix \ref{app:cu} for details. Detailed technical specifications are reported in Appendix \ref{app: implement}.
\subsection{Human Evaluation Setup}
We recruit 10 annotators and assign each 10 question–answer pairs through our dedicated annotation platform. To ensure reliability and measure inter-annotator agreement, every question is independently evaluated by two annotators. In total, the study covers 50 unique question–answer pairs. The annotation guidelines, platform details, and interface screenshots are provided in Appendix~\ref{app:human}.

\section{Results}
We present results based on our proposed \ourmodel framework and our dataset across multiple LLMs.
% \sm{two things: 1. try to add a conclusion (what does your results say) in all tables with 1-2 lines; 2. try to give reasoning why wherever applicable. Just stating the results are boring.}
% \sm{the next sentence is not needed - may you can state some objectives of the experiments here in case space allows}The analysis focuses on specificity, robustness, answer relevance and context utilization under different evaluator settings.

%################################
\subsection{Evaluation with SCORE Metrics}
\subsubsection{Specificity}
% To assess answer specificity, we check if the responses include precise, verifiable details that are directly supported by the provided documents. 
Table \ref{tab:specificity} shows the specificity scores across generator-evaluator pairs (we use $k=3$ evaluator agents). Across all settings, GPT-4o answers achieve higher specificity scores than Gemini, regardless of the evaluator. We also observe that specificity exhibits higher variance than other evaluation metrics. The higher variance in specificity scores suggests that model behavior is not uniform across questions. While GPT-4o often produces highly detailed answers, it occasionally generates less precise responses depending on the query. This variability reinforces the motivation for fine-grained specificity evaluation, as such detailed information is inherently difficult for models to provide accurately across diverse queries.
\subsubsection{Robustness}
We evaluate robustness under paraphrasing invariance and semantic perturbations. As shown in Table \ref{tab:specificity}, both GPT-4o and Gemini exhibit high robustness to paraphrasing across all evaluator settings, which indicates that superficial linguistic changes do not significantly affect model outputs.

In contrast, semantic perturbations that alter key attributes such as hazard type or location lead to noticeable consistency drops. This behavior is desirable, as it indicates that models appropriately adjust their responses when the underlying facts change. Gemini shows higher perturbation robustness scores, which suggests that its responses change less under semantic modifications. This likely reflects a tendency toward more generic answers, which remain broadly applicable even when critical details are altered, whereas GPT-4o produces more detail-sensitive responses.
\subsubsection{Answer Relevance}
To evaluate how well responses address user queries, we use a dual strategy combining masked answer relevance and BGE reranking. As shown in Table \ref{tab:specificity},  GPT-4o scored between 0.69 -- 0.73, while Gemini 2.5 scores between 0.68 -- 0.83. Additionally, using BGE reranker, we measure the value of each individual point in an answer by seeing how much the total score drops when that point is removed. Table \ref{tab:bge} shows that Gemini achieves higher relevance under this metric. We also observe frequent reordering of answer points after reranking. It indicates that the relevance model meaningfully reweights the importance of individual answer points. 

Higher relevance scores do not necessarily imply better answer quality: relevance captures semantic alignment with the query, whereas specificity reflects the presence of specific details. This motivates evaluating both metrics jointly.
% While Gemini achieves higher answer relevance scores under BGE reranking, this does not necessarily imply better answer quality.\hhs{Should we keep next two lines or not?} BGE reranking prioritizes content that is more semantically aligned with the user query, which can improve overall coherence and user interpretability of the response. In contrast, specificity captures whether precise and decision-critical details are provided. Together, these results highlight the complementary roles of relevance and specificity, and motivate their joint evaluation rather than treating either metric in isolation.
\begin{table}[h]
\centering
\small
\begin{tabular}{lcc}
\toprule
\textbf{Metric} & \textbf{GPT-4o} & \textbf{Gemini} \\
\midrule
Average BGE score & 6.69 & 7.08\\
Has answer changed (yes) & 1393 & 1397\\
\bottomrule
\end{tabular}
\caption{Comparison of BGE reranking for answers generated by GPT-4o and Gemini.}
\label{tab:bge}
 \vspace{-10pt}
\end{table}

%################################
\subsubsection{Context-Utilization}
% \paragraph{Sensitivity Analysis Across Proxy Models.} 
To test whether CU scores are sensitive to the choice of proxy model or the confidence measures, we report a sensitivity analysis in Table \ref{tab:cu_sensitivity_main}. LLaMA and Qwen variants exhibit consistent answer confidence when conditioned on retrieved documents $D$ vs. claims $C$ derived from it. This reflects consistency in the estimates as the claims $C$ are simply a summary of $D$. In contrast, Gemma and Ministral show less reliable correlations. An extended analysis is reported in Appendix \ref{app:sensitivity_cu} where LLaMA and Qwen variants also show near-perfect agreement in their CU scores and low agreement  with Gemma and Ministral. Additionally, length-normalized geometric mean of token probabilities provides consistent confidence estimates than predictive entropy across models and generators (Appendix Table \ref{tab:cu_sensitivity}). 
\begin{table}[htbp]
        \centering
        \resizebox{\columnwidth}{!}{
        \begin{tabular}{c c c}
        \toprule
                  \textbf{Evaluator}&\multicolumn{2}{c}{ $\rho( f_\theta(a|q \cup C), f_\theta(a|q \cup D)$} \\
              \cmidrule{2-3}
              &Generator=GPT-4o&Generator=Gemini2.5\\
               
              \midrule
              \textbf{LlaMA3.1-8B-Instruct} &0.83 (0.0) & 0.84 (0.0)\\
             \textbf{Gemma2-9B-it}&0.17(4.4e-11)&0.12 (1.2e-05)\\
             \textbf{Ministral-8B-Instruct} &0.14 (1.7e-07)&0.13 (6.0e-07)\\
             \textbf{Qwen2.5-7B-Instruct}&0.95 (0.0) &0.96 (0.0)\\
             \textbf{Qwen3-8B} &0.94 (0.0)&0.93 (0.0)\\
             \bottomrule
        \end{tabular}
        }
        \caption{Correlation between answer confidence conditioned on query $q$ and all claims $C$ vs. $q$ and all documents $D$ across evaluator models (p-values are reported in parenthesis). Confidence $f_\theta$ is measured in terms of geometric mean of token probabilities. Correlations for Gemma and Ministral are insignificant, indicating their insuitability for $CU$ evaluation. In contrast, LlaMA and Qwen have significantly high positive correlation.}
        \label{tab:cu_sensitivity_main}
         \vspace{-10pt}
    \end{table}
Therefore, in Table \ref{tab:cu-score} we report results with LLaMa and Qwen as evaluators using geometric-mean-based interpretation of $f_\theta$. Both $CU$ and $CU_{rel}$ (in terms of percentage drop) scores are reported. The mean $CU$ values are small and positive, indicating that, on average, removing individual claims leads to a modest but consistent drop in answer confidence. Negative minimum values reflect cases where removing a claim increases confidence, indicating distracting or redundant context\textemdash a behavior expected in realistic retrieval settings. Importantly, these negatives are rare and do not affect the overall positive mean, implying that most retrieved claims contribute constructively. Relative CU values provide a more interpretable signal. Mean $CU_{rel}$ values indicate that on average, removing a single claim reduces answer confidence by 1–2\%, with a small subset of claims causing substantially larger drops (9-10\%), indicating concentrated evidence dependence amid otherwise distributed evidence usage. Large negative values (e.g., -8.56\%) highlight claims that harm generation. This demonstrates that the measure is effective at surfacing both highly supportive and detrimental evidence.
    \begin{table*}[htbp]
        \centering
        \begin{tabular}{l c ccc ccc}
        \toprule
\textbf{Generator}&\textbf{Evaluator} &\multicolumn{3}{c}{\textbf{ $CU$}}& \multicolumn{3}{c}{ $CU_{rel}^\%$  } \\
             \cmidrule{3-8}
             & &min&max&mean &min&max&mean\\
             \midrule
\multirow{3}{*}{GPT-4o}&LLaMA3.1-8B-Instruct&  -0.001 &0.024& 0.004 (0.002)& -0.17& 9.82& 1.63 (0.87)  \\
% &Qwen2-7B&-0.001& 0.025 &0.005 (0.003)&   -0.33& 10.93& 1.90 (1.03)\\
&Qwen3-8B &0.000&    0.019&0.004 (0.002)&-0.51&9.63&1.74 (0.94)\\
\midrule
\multirow{3}{*}{Gemini2.5}&LLaMA3.1-8B-Instruct& -0.001& 0.014& 0.003(0.002) & -8.56& 6.45&1.34 (0.83)\\
% &Qwen2-7B& -0.002&  0.014&0.003 (0.002)& -0.98& 8.33&1.52(0.94)\\
&Qwen3-8B& -0.001&0.015 &0.003 (0.002)&-0.74& 7.35&1.39 (0.87)\\
\bottomrule
        \end{tabular}
        \caption{$CU$ scores across two generators and two evaluators (standard deviation is in parentheses), showing on average modest but positive contributions from individual claims to answer confidence, with rare distracting claims. Relatively larger max values indicate concentrated evidence dependence.}
         \vspace{-10pt}
        \label{tab:cu-score}
    \end{table*}
%################################
\subsection{Human Evaluation}
We conduct a human evaluation comprising two analyses. First, we measure inter-annotator agreement to assess the consistency of human labels. Second, we show the agreement between human labels and automated evaluators (GPT and Qwen) to understand how the model-based scores aligns with human judgment. Given the open-ended and domain-specific nature of the task, some disagreement is expected, particularly for attributes that require interpretation rather than direct verification.
\noindent\paragraph{Inter-Annotator Agreement. }We compute inter-annotator agreement across the 50 double-annotated questions. Table \ref{tab:human_annotation} shows agreement percentages (see Table \ref{tab:human_annotation1} for details). Agreement is higher for concrete attributes such as hazard and location, and lower for subjective dimensions such as timeline, which often requires interpretation of vague temporal statements. For continuous metrics, we evaluate agreement using Spearman correlation. We obtain a Spearman correlation of 0.22 for answer relevance and 0.35 for context utilization, highlighting that both tasks are inherently subjective and cognitively demanding, with context utilization being slightly more stable than relevance.

\noindent\paragraph{Human vs. Automated Agreement.}
We compare human annotations with two automatic evaluators using row-level exact match on the same label ranges. Table~\ref{tab:human_annotation} shows that alignment is generally higher for hazard and location, but drops for timeline and intensity (see Table \ref{tab:human_annotation2} for details). We also notice that many human--automated mismatches occur on the same questions where annotators themselves disagree, which suggests that a large fraction of the errors come from genuinely ambiguous or subjective cases. For answer relevance, the Spearman correlations are 0.33 (human vs. GPT) and 0.32 (human vs. Qwen); for context utilization, they are 0.43 (for both human vs. Qwen and human vs. LLaMA). Matching or exceeding human–human agreement, our automated metrics appear as reasonable human-aligned proxies.
\paragraph{Error Analysis }We conduct a qualitative error analysis to understand common failure modes and sources of disagreement. For specificity-hazard, some answers discuss multiple hazards (e.g., ice storms, storm surge, sea level rise) in the same response. In these cases, some annotators score hazard specificity only with respect to the hazard explicitly asked in the question, while others treat any hazard details mentioned in the answer as sufficient. For specificity-timeline, we observe boundary cases where the answer mentions a timeline and the retrieved sources also include temporal information, but the timelines do not match. In one example, the answer claims impacts over the ``next 30--50 years,'' but that exact range is not stated in the retrieved evidence. The human annotators  marked \emph{Yes} because a timeline was mentioned, GPT marked \emph{N/A} due to vague temporal framing, and Qwen marked \emph{No} because the ``30--50 years'' claim was not verifiable from the sources. Additional examples are provided in Appendix~\ref{app:human}.
\begin{table}
\centering
\resizebox{0.95\columnwidth}{!}{
\begin{tabular}{@{}lccc@{}}
\toprule
\textbf{Metric} & \textbf{Annotators} & \textbf{Anno-GPT} & \textbf{Anno-Qwen} \\
\midrule
\textbf{Hazard}     &  92 & 88 & 88      \\
\textbf{Location}  &   86&77& 73 \\
\textbf{Timeline}  &  52 &45 & 40\\
\textbf{Intensity} &  78 &75& 59\\
\bottomrule
\end{tabular}
}
\caption{Agreement percentage between annotators, annotator and automated metrics for specificity.}
 \vspace{-10pt}
\label{tab:human_annotation}
\end{table}
% \begin{table}[ht]
% \centering

% \resizebox{0.95\columnwidth}{!}{
% \begin{tabular}{@{}lcc@{}}
% \toprule
% \textbf{Metric} & \textbf{Good Answer Score} & \textbf{Corrupted Answer Score} \\
% \midrule
% RAGAS (Relevance)         &        &        \\
% \textbf{Ours: Relevance}         &        &        \\
% \bottomrule
% \end{tabular}
% }
% \caption{Contrastive Evaluation: Metric scores on good vs. corrupted answers. A good metric should assign a higher score to the unperturbed (correct) version.}
% \label{tab:ablation}
% \end{table}

% \clearpage
% \input{latex/5_Results}
% \clearpage
\section{Conclusion}
We introduce \ourmodel, a multi-dimensional evaluation framework for assessing LLM-generated answers in domain-specific, high-stakes settings, together with a new dataset designed to support evaluation where gold-standard references are scarce. By jointly analyzing specificity, robustness, answer relevance, and context utilization, we show that models exhibit distinct trade-offs between producing detailed, precise information and generating broadly relevant but more generic responses. Our human evaluation highlights the challenges of aligning automated metrics with human judgment in open-ended, expert-level tasks, particularly for attributes requiring significant interpretation. These results emphasize the importance of evaluating LLM outputs using complementary metrics rather than a single score for a more reliable assessment of LLMs deployed in real-world decision support.

\clearpage
\section*{Ethical Considerations}

This work focuses on the evaluation of large language model (LLM) outputs rather than the deployment of automated decision-making systems. Although our framework targets high-stakes domains such as natural hazard response and infrastructure planning, we emphasize that such systems should complement rather than replace human expert judgment, particularly in decisions affecting public safety and critical infrastructure. Our dataset is synthetically generated and contains no personal or personally identifiable information; however, while it is designed to reflect realistic hazard–location mappings, it may not capture the full complexity and diversity of real-world hazard scenarios, potentially disadvantaging regions or communities underrepresented in the climate and infrastructure literature. We also acknowledge that evaluation metrics themselves can introduce ethical risks if misinterpreted, as high scores on relevance, robustness, or specificity may create false confidence if treated as definitive indicators of system safety or deployment readiness rather than complementary assessment tools. Finally, all human annotators involved in this study were recruited and treated in accordance with institutional review board (IRB) guidelines, provided with clear information about the research purpose, participated voluntarily, and were compensated fairly for their time.
\section*{Limitations}

The current framework relies on a specific knowledge base of approximately 600,000 climate-related records. A primary limitation is that if this knowledge base is insufficient or lacks coverage for a specific niche, the system may retrieve irrelevant documents. This can lead to error propagation where the model generates an answer based on poor grounding. In addition, human evaluation of open-ended, domain-specific questions exhibits inherent subjectivity, particularly for attributes that requires interpretation such as timeline and intensity.

% \section*{Acknowledgments}
% \todo{need to add funding info.};;; \sm{DO NOT have this for the submission; add after it is accepted}
\section*{Acknowledgment}
We thank Sourav Medya for his valuable feedback on the manuscript. This work was supported by UChicago Argonne, LLC, Operator of Argonne National Laboratory (“Argonne”). Argonne’s work was supported by the U.S. Department of Energy, Grid Deployment Office, under contract DE-AC02-06CH11357. This research used resources of the Argonne Leadership Computing Facility at Argonne National Laboratory, which is supported by the Office of Science of the U.S. Department of Energy, Office of Science, under contract number DE-AC02-06CH11357.

\bibliography{custom}
\clearpage

\clearpage
\appendix
\section{Dataset Details}
\label{app:data}
Here we provide details of the dataset generation.

\begin{table*}[htbp]
% \centering
\centering
\small
\resizebox{\textwidth}{!}{
\begin{tabular}{p{2cm} p{12cm}}
\toprule
\textbf{Sector} & \textbf{Profession} \\
\midrule

Transportation &
Highway Engineer, Bridge Inspector, Railway Systems Engineer, Transit Operations Manager, Airport Infrastructure Manager, Port Facility Manager, Transportation Safety Inspector, Traffic Systems Engineer, Pavement Engineer, Transportation Planner \\[4pt]

Water &
Water Systems Engineer, Hydraulic Engineer, Dam Safety Inspector, Wastewater Treatment Specialist, Maritime Infrastructure Manager, Stormwater Engineer, Water Quality Specialist, Coastal Infrastructure Engineer \\[4pt]

Energy &
Power Systems Engineer, Electrical Grid Manager; Energy Distribution Specialist, EV Infrastructure Planner, Renewable Energy Systems Manager, Transmission Line Engineer, Substation Engineer, Energy Storage Specialist \\[4pt]

Buildings &
Structural Engineer, Building Systems Manager, Facilities Manager, Real Estate Asset Manager, Building Automation Specialist, Construction Manager, Building Code Inspector, MEP Systems Engineer \\[4pt]

Communications &
Telecommunications Engineer, Broadband Infrastructure Specialist, Network Resilience Manager, Data Center Infrastructure Engineer, Fiber Optics Specialist, Communications Systems Planner, Network Security Engineer \\
\bottomrule

\end{tabular}
}
\caption{List of professions across five critical infrastructure sectors (Transportation, Water, Energy, Buildings, and Communications) used for generating synthetic user profiles in our dataset.}
\label{tab:profession}
\end{table*}

\subsection{Profession}
Table \ref{tab:profession} provides a summary 40  different professions across five sectors.

\subsection{Hazard and Location}
Table \ref{tab:hazard_location} summarizes the seven hazard types and their corresponding locations (counties and states) used in the question generation template.
\begin{table*}[ht]

\centering
\small
\resizebox{\textwidth}{!}{
\begin{tabular}{p{1.5cm}p{13cm}}
\toprule
\textbf{Hazard Type} & \textbf{Counties and States} \\
\midrule
Coastal Flooding &
Bergen, NJ; Atlantic, NJ; Ocean, NJ; Cape May, NJ; Hudson, NJ; Monmouth, NJ; Grays Harbor, WA; Middlesex, NJ; Kings, NY; Cumberland, NJ; Clatsop, OR; Cameron, TX; Philadelphia, PA; Queens, NY; Coos, OR; Bronx, NY; Sussex, DE; Westchester, NY; New York, NY; Jefferson, LA; Fairfield, CT; St. Charles, LA; Suffolk, NY; Aransas, TX; Union, NJ
\\ 
\midrule

Cold Wave &
Cook, IL; Milwaukee, WI; Minnehaha, SD; Wayne, MI; Lake, IL; Nueces, TX; Lake, IN; Hennepin, MN; Williams, ND; Will, IL; Yakima, WA; Anoka, MN; Flathead, MT; Winnebago, IL; Pennington, SD; Dane, WI; Ramsey, MN; Cass, ND; Marathon, WI; Sheboygan, WI; Blue Earth, MN; Brown, WI; Olmsted, MN; Outagamie, WI; St. Louis, MN
\\ 
\midrule

Drought &
Santa Barbara, CA; Yolo, CA; Sutter, CA; Napa, CA; Colusa, CA; Glenn, CA; Butte, CA; Sonoma, CA; Sacramento, CA; Solano, CA; Pinal, AZ; Floyd, TX; Lubbock, TX; Humboldt, NV; Doña Ana, NM; Maricopa, AZ; Yuma, AZ; Kings, CA; Imperial, CA; Merced, CA; Madera, CA; Stanislaus, CA; Fresno, CA; Tulare, CA; Kern, CA
\\
\midrule

Heat Wave &
Cook, IL; Clark, NV; St. Louis, MO; Philadelphia, PA; Dallas, TX; Tulsa, OK; Maricopa, AZ; Queens, NY; Tarrant, TX; Kings, NY; Oklahoma, OK; Tulare, CA; Jackson, MO; Shelby, TN; Baltimore, MD; Fulton, GA; Los Angeles, CA; Harris, TX; Bexar, TX; Fairfax, VA; Franklin, OH; DeKalb, GA; Prince George's, MD; Mecklenburg, NC; Wayne, MI
\\ 
\midrule

Hurricane &
Harris, TX; Miami-Dade, FL; Broward, FL; Palm Beach, FL; Hillsborough, FL; Lee, FL; Brevard, FL; Pinellas, FL; Charleston, SC; Pasco, FL; Horry, SC; Collier, FL; Chatham, GA; Mobile, AL; New Hanover, NC; Galveston, TX; Orange, FL; Volusia, FL; Indian River, FL; St. Lucie, FL; St. Johns, FL; Manatee, FL; Clay, FL; Beaufort, SC; Escambia, FL
\\ 
\midrule

Ice Storm &
Nassau, NY; Tulsa, OK; Greene, MO; Lancaster, NE; St. Louis, MO; Oakland, MI; Boone, MO; Richland, SC; Monmouth, NJ; Washington, AR; Macomb, MI; Johnson, KS; Morris, TX; Baxter, AR; Rogers, OK; Douglas, NE; Sedgwick, KS; Linn, IA; Dubuque, IA; Stark, OH; Polk, IA; Peoria, IL; Knox, TN; Lucas, OH; Hamilton, OH \\ 
\midrule

Wildfire &
San Diego, CA; Riverside, CA; San Bernardino, CA; Los Angeles, CA; Washington, UT; Elko, NV; Ventura, CA; Orange, CA; Pima, AZ; Maricopa, AZ; Ravalli, MT; Kern, CA; Yavapai, AZ; Utah, UT; Madera, CA; Nevada, CA; Placer, CA; Shasta, CA; Siskiyou, CA; Tehama, CA; Santa Cruz, CA; Alameda, CA; Tuolumne, CA \\

\bottomrule

\end{tabular}
}

\caption{Hazard types and the corresponding counties and states included in our dataset generation framework.}
\label{tab:hazard_location}
% \vspace{-1mm}
\end{table*}

% \FloatBarrier
% \clearpage
\section{Evaluation Framework}
\label{app:method}
In this section, we provide additional details on our evaluation framework.
\subsection{Specificity}
\label{app:spec}
\textbf{Atomic Claim Decomposition.} We begin by decomposing each generated answer into a set of atomic claims, where each claim captures a single, self-contained factual statement. This decomposition enables us to evaluate the output at a more granular level, rather than treating the entire answer as one unit. By focusing on individual claims, we can perform more precise and targeted assessments of specificity. We can also identify whether each claim includes the fine-grained details that make it informative.\\
\textbf{Specific Information Extraction.}
After decomposing each generated answer into claims, we extract the specific information that can be verified against the retrieved documents. In particular, we focus on four dimensions of specificity: \textit{hazard type}, \textit{location}, \textit{timeline}, and \textit{intensity}. To accomplish this, we prompt an LLM to identify and extract all explicit mentions of these dimensions from the claims. The extracted information is then passed to our specificity evaluation agents in the following step for verification and scoring.\\
\textbf{Specificity Scoring. }We evaluate each claim to determine whether it includes specific factual details that are supported by the knowledge base. The goal of this step is not to assess whether a claim is factually correct, but to determine whether fine grained information such as hazard type, location, timeline, or intensity is verified based on the knowledge base. We denote the input to the evaluation process as a triplet:$(S, c_i, K)$, where $S$ is specific information, $c_i$ is the $i$-th atomic claim extracted from the generated answer, K is the corresponding knowledge base (e.g., abstracts from the literature). This triplet is provided to a language model evaluator agent, which is prompted to assess whether the claim \( c_i \) expresses concrete information that is \textit{verifiable using the knowledge base} \( K \), with respect to four pre-defined specificity dimensions:

$\mathcal{D} = \{\text{hazard type},\ \text{location},\ \text{timeline},\ \text{intensity}\}$
Each agent produces a structured annotation for the claim across the four specificity dimensions from a fixed label set: $L=\{yes, no, n/a\}$. A label of “yes” indicates that the specific information is present in the claim and is also verifiable using the knowledge base. A label of “no” indicates that the information is not supported by the knowledge. Finally, “n/a” is used when a particular dimension does not apply to the claim’s context (e.g., that specific information is not mentioned in the claim).

% \clearpage
% \FloatBarrier
% \subsection{Answer Relevance}
% \label{app:are}
% BGE reranking scores are shown in Table \ref{tab:bge}.
% \begin{table*}[htbp]
% \centering
% \small
% \begin{tabular}{lcc}
% \toprule
% \textbf{Metric} & \textbf{GPT-4o} & \textbf{Gemini} \\
% \midrule
% Average BGE score & 6.69 & 7.08\\
% Has answer changed (yes) & 1393 & 1397\\

% \bottomrule
% \end{tabular}
% \caption{Comparison of BGE reranking for answers generated by GPT-4o and Gemini.}
% \label{tab:bge}
% \end{table*}

\subsection{Context Utilization: Proxy-Based Scoring}
\label{app:cu}
We operate at the level of atomic claims rather than retrieved chunks. Claims are extracted from retrieved context using GPT-4o, prompted to identify unique, non-redundant factual propositions. This claim-level representation enables fine-grained attribution while avoiding redundancy within retrieved passages. To quantify each claim's contribution, we require an estimate of the model's confidence in generating a fixed answer given a particular set of claims. However, answers are often produced by closed or proprietary models (e.g., GPT or Gemini APIs) that do not expose token-level likelihoods. To enable reproducible and model-agnostic evaluation, we therefore decouple answer generation from confidence estimation. Concretely, for a fixed answer $a$, we compute confidence using teacher forcing on a lightweight open-source proxy language model. The proxy model is conditioned on the query $q$ and context $X$, and token-level probabilities are estimated by forcing the model to follow the reference answer sequence. This yields a consistent likelihood estimate without requiring access to the original model's internal probabilities.
\paragraph{Sensitivity Analysis of Proxy Models.}
\label{app:sensitivity_cu}A natural concern is whether the CU scores are sensitive to the choice of proxy models or confidence measures. To address this, we conduct a sensitivity analysis and measure:
\begin{enumerate}
    \item \textbf{Confidence Consistency} as the correlation $\rho( f_\theta(a|q \cup C),  f_\theta(a|q \cup D))$ between answer confidence given query $q$ and all claims $C$ and $q$ and the original retrieved documents $D$.
    \item \textbf{CU Agreement}: correlation in CU scores across proxy models.
\end{enumerate}
Table \ref{tab:cu_sensitivity} reports Confidence Consistency---correlation between answer confidence when conditioned on query $q$ and claims $C$ ($f_\theta(a|q \cup C)$) extracted from retrieved information $D$ compared to conditioning on query and $D$. These are computed based on both our confidence measures\textemdash geometric mean of token-level probabilities and predictive entropy. If a proxy model’s token-level probabilities are well calibrated, the two would be highly correlated, since $C$ summarizes $D$. LLaMA and Qwen variants exhibit high confidence consistency, while Gemma and Ministral show substantially lower correlations, suggesting they might be less suitable for proxy-based CU estimation. Across models, normalized log-likelihood yields more consistent confidence estimates than predictive entropy. 

Further, Table \ref{tab:cu_sensitivity_modelpair} reports \textbf{CU Agreement} as the pairwise correlation in $CU$ and $CU_{\text{rel}}$ across proxy models. CU scores from LLaMA and Qwen variants are nearly perfectly correlated, whereas correlations involving Gemma and Ministral are lower, although they remain positive. These patterns are consistent for answers generated by both GPT-4o and Gemini-2.5.

\begin{table*}[htbp]
        \centering
        \resizebox{.7\textwidth}{!}{
        \begin{tabular}{c c c c c}
        \toprule
              \textbf{Evaluator}& \multicolumn{2}{c}{Generator=GPT-4o}&\multicolumn{2}{c}{Generator=Gemini2.5}\\
              \cmidrule{2-5}
              &( $f_\theta$=log-prob) &($ f_\theta$=entropy)&($ f_\theta$=log-prob) &($ f_\theta$=entropy) \\
              \midrule
              \textbf{LlaMA3.1-8B-Instruct} &0.83 (0.0) &0.40 (5.7e-55) & 0.84 (0.0)&0.58 (2.1e-128)\\
             \textbf{Gemma2-9B-it}&0.17(4.4e-11)&0.16(4.2e-09)&0.12 (1.2e-05)&0.19 (2.1e-13) \\
             \textbf{Ministral-8B-Instruct} &0.14 (1.7e-07)&0.13 (2.0e-06)&0.13 (6.0e-07)&0.38 (3.2e-50)\\
             \textbf{Qwen2.5-7B-Instruct}&0.95 (0.0) &0.46 (3.9e-76)&0.96 (0.0)&0.62 (5.8e-150)\\
             \textbf{Qwen3-8B} &0.94 (0.0)&0.49 (1.7e-85)&0.93 (0.0)&0.60 (5.3e-137)\\
             \bottomrule
        \end{tabular}
        }
        \caption{Correlation ($\rho$) between $ f_\theta(a|q \cup C)$ and $ f_\theta(a|q \cup D)$. Where $C$ denotes all extracted claims from retrieved content $R$. P-values are reported in parentheses. The correlations for Gemma and Ministral are positive but low, indicating the token probabilities might not be well-calibrated.}
        \label{tab:cu_sensitivity}
    \end{table*}
   \begin{table*}[htbp]
        \centering
        \resizebox{0.7\textwidth}{!}{
        \begin{tabular}{c c c c c}
        \toprule
             \multirow{2}{*}{\textbf{Evaluators}}&\multicolumn{2}{c}{Generator=GPT-4o}&\multicolumn{2}{c}{Generator=Gemini2.5}\\
              \cmidrule{2-5}
              &$\Delta$&$\delta$&$\Delta$&$\delta$\\
              \midrule 
              \textbf{LLaMA vs. Gemma}&-0.00 (0.87)&-0.09 (0.0)& -0.10 (2.94e-04)&-0.09 (3.87e-04)\\
            \textbf{Ministral vs. Gemma} &0.66 (1.2e-177)&0.01 (0.63)&0.45 (1.2e-69)& 0.01 (0.68)\\
              \textbf{Ministral vs. LLaMA} &0.03 (0.25)&0.06 (0.02)& -0.02 (0.48)&0.01 (0.69)\\
              \textbf{Ministral vs. Qwen2.5}&0.00 (0.87)&0.06 (0.04)&-0.01 (0.7)& 0.05 (0.06)\\
              \textbf{Qwen2.5 vs. LLaMA}&0.87 (0.0)&0.86 (0.0)& 0.84 (0.0)& 0.80 (2.5e-321)\\
              \textbf{Qwen2.5 vs. Gemma} &-0.02 (0.42)&0.01 (0.77)& -0.10 (1.7e-04)& -0.04 (.13)\\
            \textbf{Ministral vs. Qwen3}&-0.03 ( 0.32)&0.05 (0.04)& -0.02 (0.41)& 0.02 (0.43) \\
            \textbf{Qwen3 vs. LLaMA}&0.81 ( 0.0)&0.83 (0.0)&0.85 (0.0)& 0.79 (1.2e-295)\\
            \textbf{Qwen3 vs. Gemma }& -0.03 (0.30)&0.00 (1.00)&-0.08 (.002)&-0.03 (0.2)\\
             \textbf{Qwen3 vs. Qwen2.5} & 0.82 (0.0)&0.83 (0.0)&0.83 (0.0)&0.82 (0.0)\\
             \bottomrule
        \end{tabular}
        }
        \caption{Pairwise correlation between context utilization scores from different proxy models using both absolute ($\Delta$) and relative ($\delta$) scores. P-values are reported in parentheses.}
        \label{tab:cu_sensitivity_modelpair}
    \end{table*}
\subsection{Readability}
\label{app:readability}

The next metric we show is readability. Since our questions originate from domain-specific, graduate-level, or professional contexts, we expect the model-generated answers to match an appropriate level of linguistic complexity. To evaluate this, we employ two widely used readability measures. The first is the Flesch Reading Ease (FRE) score \cite{flesch1948new}, a numerical metric ranging from 0 to 100+, where higher values indicate easier text. The FRE score is defined as:
\begin{equation}
\text{FRE} = 206.835 - 1.015 \cdot \text{ASL} - 84.6 \cdot \text{ASW}
\end{equation}
where $\text{ASL}$ denotes the average sentence length (words per sentence), and $\text{ASW}$ represents the average number of syllables per word. According to conventional interpretation, scores above 60 correspond to plain English that is understandable to the general public (approximately 8th--9th grade reading level), while scores below 30 indicate highly complex or academic text, typically suitable for college-level readers or above.

The second metric we consider is the Flesch–Kincaid Grade Level (FKGL) \cite{kincaid1975derivation}, which maps text complexity directly onto a U.S. school grade level. FKGL is defined as:
\begin{equation}
\text{FKGL} = 0.39 \cdot \text{ASL} + 11.8 \cdot \text{ASW} - 15.59
\end{equation}
A score of 8.0 indicates that the text is readable by an average eighth-grade student, whereas a score of 16.0 or higher suggests that graduate-level reading proficiency is required.

We apply these metrics to evaluate and compare the readability of responses generated by two prominent large language models: \textbf{GPT-4o} and \textbf{Gemini}. For each generated response, we compute the FRE and FKGL scores and categorize them into standard readability levels. These include very easy, plain English, college level, and graduate level. Similarly, FKGL scores are grouped into categories such as elementary, middle school, high school, undergraduate, and graduate reading levels.

A summary of this analysis is presented in Table~\ref{tab:readability_summary}. This includes average FRE and FKGL scores for each model, as well as the distribution of their outputs across interpretive readability levels. These results allow us to assess not only the average complexity of model outputs, but also how frequently they produce text that is potentially inaccessible to a general audience.

\begin{table}[htbp]
\centering
\small
\begin{tabular}{p{4.2cm}cc}
\toprule
\textbf{Metric} & \textbf{GPT-4o} & \textbf{Gemini} \\
\midrule
Average Flesch Reading Ease & 12.84 & 10.98\\
Average Flesch-Kincaid Grade Level & 17.80 & 18.22\\
\midrule

\#Fairly difficult (10th–12th grade) & 0 & 2 \\
\# Difficult to read (college level) & 37& 93 \\
\# Very difficult (college graduate or higher) & 1375 & 1317 \\
\midrule
\#Middle school level  & 1 & 2 \\
\#High school level & 0 & 28 \\
\# College undergraduate level & 167 & 295 \\
\# Graduate/professional level & 1244 & 1087 \\
\bottomrule
\end{tabular}
\caption{Comparison of automated readability scores for answers generated by GPT-4o and Gemini.}
\label{tab:readability_summary}
\end{table}
\section{Human Evaluation}
\label{app:human}
\begin{figure*}[t]
    \centering
    \includegraphics[width=0.98\textwidth]{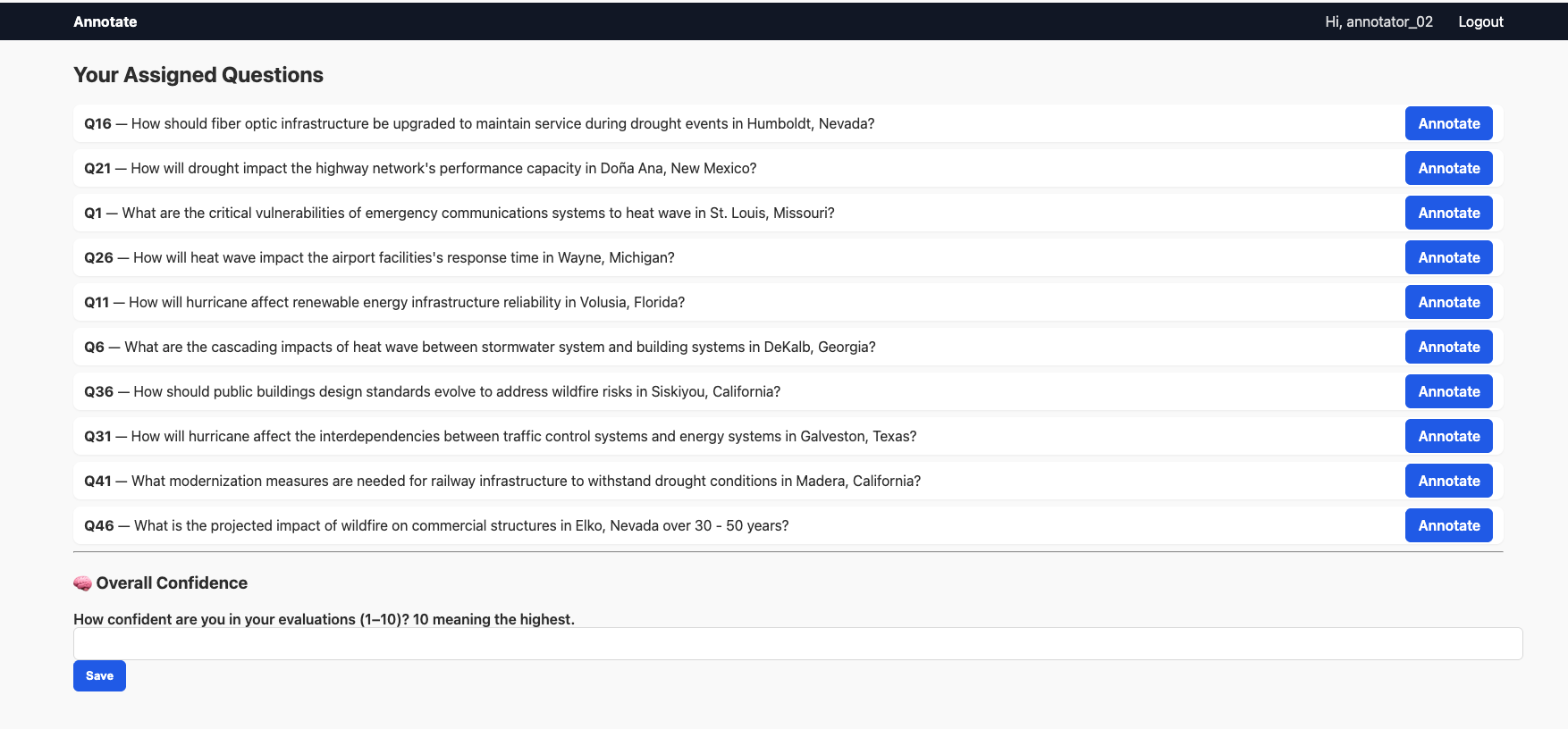}
    \caption{The annotation dashboard where each annotator receives ten questions and evaluates them by selecting the “Annotate” button for each item.}
    \label{fig:annnotation_login}
\end{figure*}

\begin{figure*}[t]
    \centering

    \begin{subfigure}{0.98\textwidth}
        \centering
        \includegraphics[width=\textwidth]{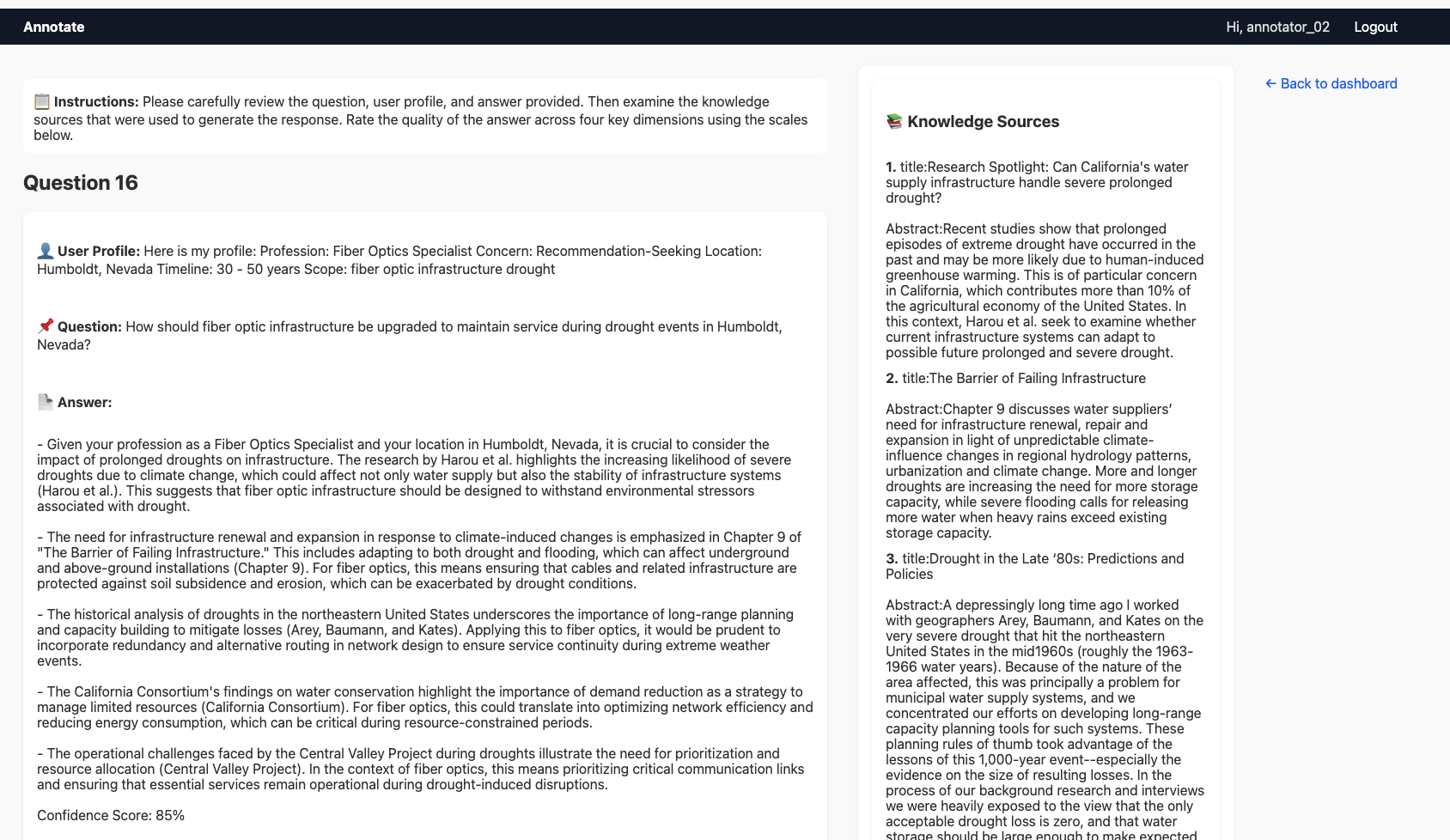}
        \caption{Annotation Interface – Part 1}
        \label{fig:annot1}
    \end{subfigure}

    \vspace{1em} 

    \begin{subfigure}{0.98\textwidth}
        \centering
        \includegraphics[width=\textwidth]{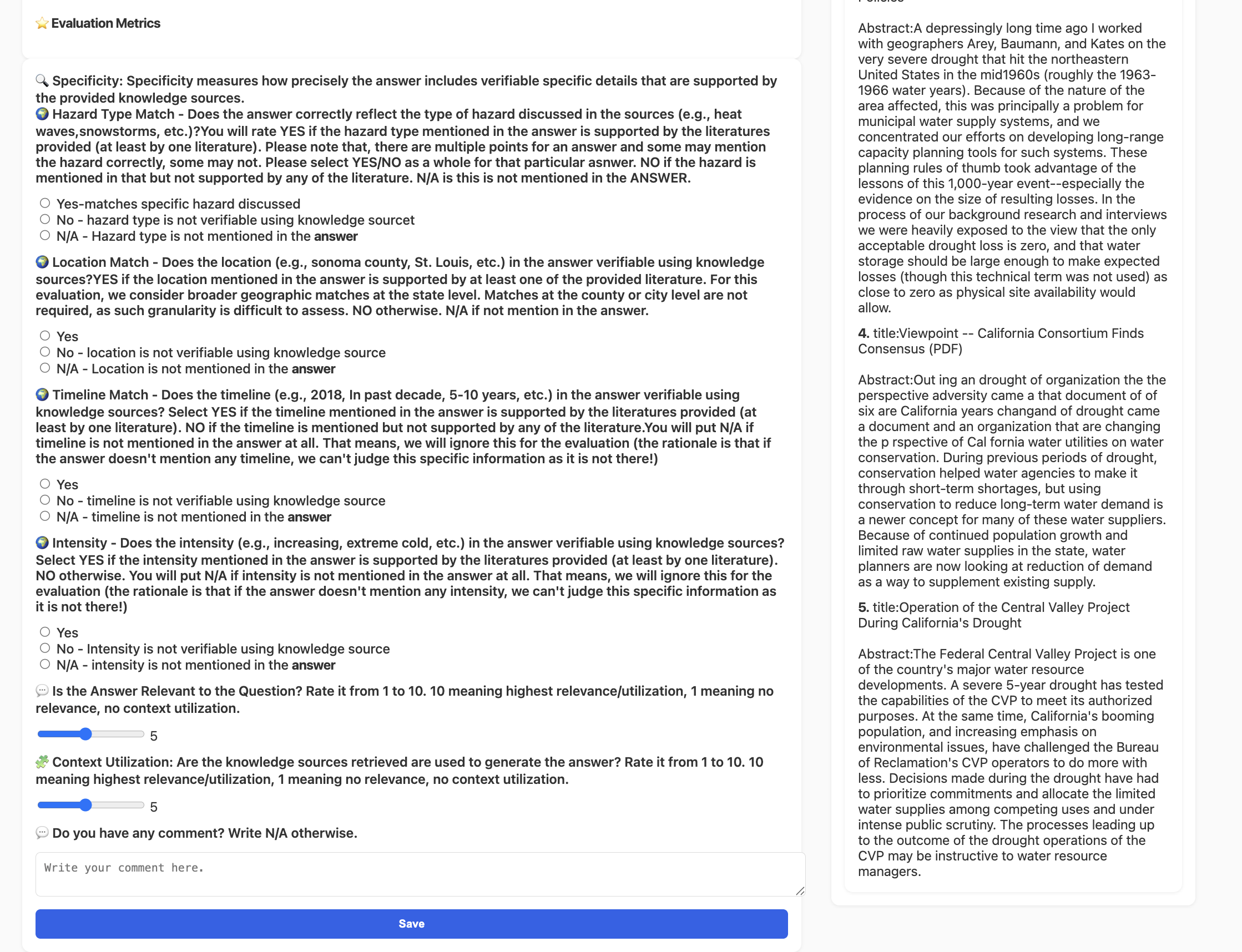}
        \caption{Annotation Interface – Part 2}
        \label{fig:annot2}
    \end{subfigure}

    \caption{Human Annotation Interface of our evaluation pipeline.}
    \label{fig:annotation}
\end{figure*}

We recruit 10 annotators to validate our evaluation scores. Each annotator first logged into the evaluation platform and was presented with a personalized dashboard listing their assigned questions (see Figure \ref{fig:annnotation_login}). The dashboard clearly enumerates each question (e.g., Q16, Q21, Q1, etc.) along with an “Annotate” button that opens the detailed evaluation page. Upon selecting a question, annotators were shown a structured evaluation interface (see Figure \ref{fig:annotation}) that displayed the user profile used to generate the answer alongside the question itself, the model-generated answer (formatted in numbered points following our prompt design), and the five retrieved knowledge sources that informed the response. Annotators then evaluated each answer across four dimensions: Specificity, which included sub-criteria for hazard type match, location match, timeline match, and intensity match; Answer Relevance, rated on a 1–10 scale; Context Utilization, which assessed whether the retrieved documents were actually used in the generated answer; and Overall Confidence, where annotators reported how confident they were in their evaluation (1–10) and could optionally provide comments. Each metric included clear guidance indicating when an annotator should select Yes, No, or N/A, along with radio-button options to standardize responses.

\noindent\textbf{Agreement. }Table \ref{tab:human_annotation1} shows agreement, disagreement and agreement percentage between human annotators over 50 double-annotated question–answer pairs. Table \ref{tab:human_annotation2} shows the same metrics for agreement between human labels and automated evaluators. Note that each question is annotated by two annotators, yielding 100 question-answer pairs.
\begin{table}[htbp]
\centering
\resizebox{0.95\columnwidth}{!}{
\begin{tabular}{@{}lcccc@{}}
\toprule
\textbf{Metric} & \textbf{Agree } & \textbf{Disagree} & \textbf{Agree (\%)} & \textbf{Fleiss’ kappa} \\
\midrule
\textbf{Hazard}              &46& 4 &92  & 0.6678 \\
\textbf{Location}           & 43 & 7 & 86 & 0.6834 \\
\textbf{Timeline}            &  26 & 24 & 52 & 0.2640 \\
\textbf{Intensity}            & 39  & 11 & 78 & 0.4579 \\
\bottomrule
\end{tabular}
}
\caption{Inter-annotator agreement across 50 double-annotated questions. Agreement is higher for more concrete attributes (e.g., hazard, location) and lower for more subjective ones (e.g., timeline).}
\label{tab:human_annotation1}
\end{table}
\begin{table}[htbp]
\centering
\resizebox{0.80\columnwidth}{!}{
\begin{tabular}{lcccccc}
\toprule
\textbf{Metric} 
& \multicolumn{2}{c}{\textbf{Agree}} 
& \multicolumn{2}{c}{\textbf{Disagree}} & \multicolumn{2}{c}{\textbf{Agree (\%) }} \\
\cmidrule(lr){2-3} \cmidrule(lr){4-5}\cmidrule(lr){6-7}
& \textbf{GPT} & \textbf{Qwen} & \textbf{GPT} & \textbf{Qwen}& \textbf{GPT} & \textbf{Qwen} \\
\midrule
Hazard  &88 &88 &12 &88 &88  &88  \\
Location   &77&73&23&27&77&73 \\
Timeline   &45&40&55&60&45& 40 \\
Intensity &75&59&25&41&75&59 \\
\bottomrule
\end{tabular}
}
\caption{Human vs. automated evaluator agreement over 100 human-annotated rows (50 questions × 2 annotators), computed as row-level exact match. Alignment is higher for hazard and location and lower for timeline, intensity, and answer relevance.}
\label{tab:human_annotation2}
\end{table}
\\
\textbf{Error Analysis.}
\begin{figure*}
    \centering
    \includegraphics[width=0.95\textwidth]{
    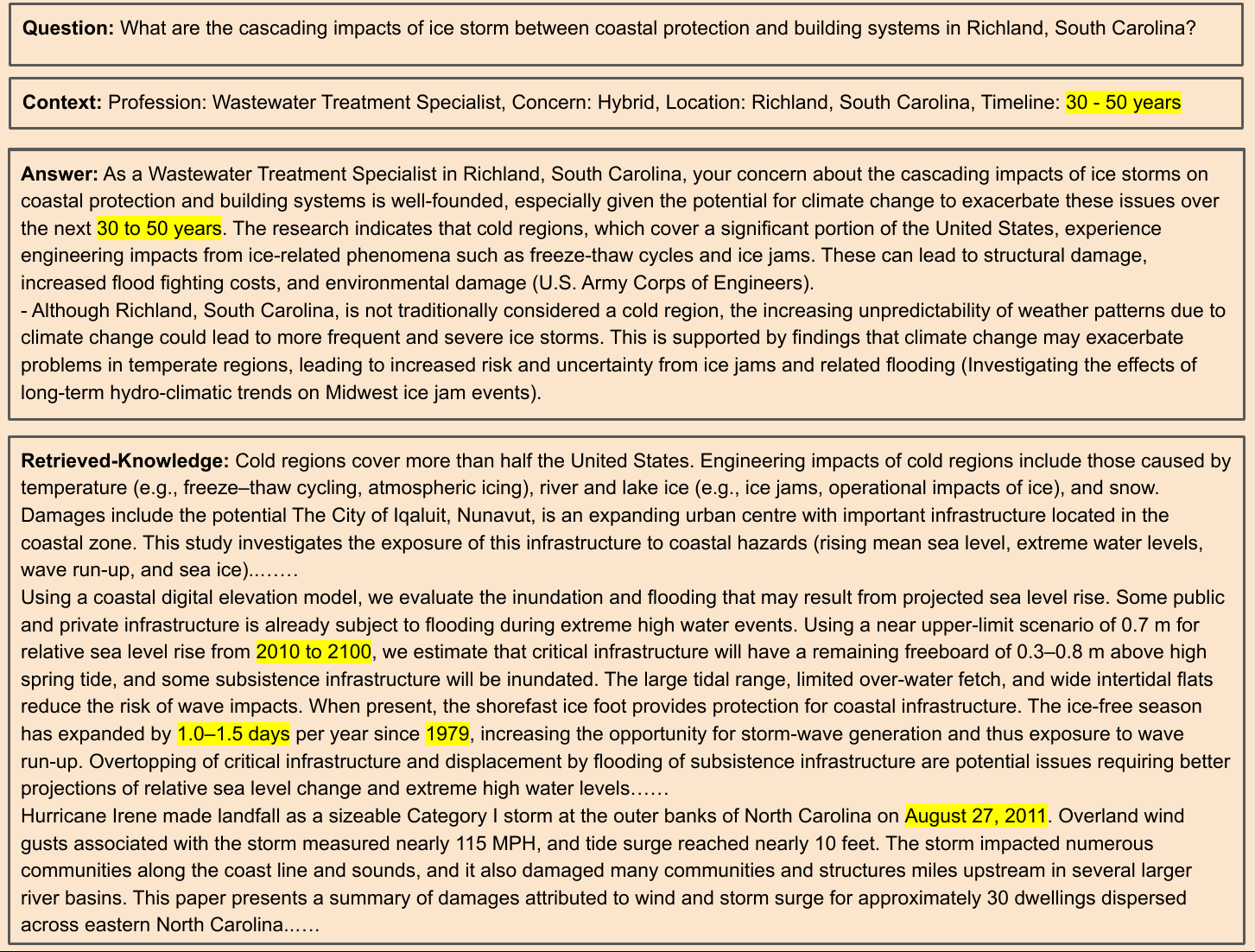}
    \caption{Example from the error analysis showing the question, context, retrieved evidence, and generated answer. The highlighted text shows temporal information.}
    \label{fig:error}
\end{figure*}
Figure \ref{fig:error} shows one representative example, including the question, context, retrieved knowledge (partial), and the generated answer. In this case, the answer includes temporal information, but it is stated in a vague form compared to the evidence. The retrieved sources discuss time-related effects, but they do not support that exact range. This creates an ambiguity: one may mark the timeline as supported because a timeline is mentioned, while a stricter interpretation marks it as unsupported because the claimed timeline is not explicitly verifiable from the provided evidence.
\section{Implementation Details}
\label{app: implement}
All experiments are conducted using fixed decoding and evaluation configurations to ensure consistency across both proprietary and open-source models. For answer generation, we use a temperature of 0.1 and a top-p of 0.9 for both GPT-4o and Gemini-2.5-flash to prioritize factual precision and minimize stochastic variation, which is particularly important in high-stakes domains.
For literature retrieval, we use a semantic search system built on a knowledge base containing approximately 600,000 climate-related records. Query embeddings are generated using the all-MiniLM-L6-v2 sentence transformer model and indexed with FAISS (Facebook AI Similarity Search). For each query, we retrieve the top five relevant literature abstracts ($k = 5$), which are used as the grounding context for all generated responses.
All proprietary models are accessed via APIs. Local model inferences for LLaMA3.1-8B, Qwen3-8B etc., are performed using the vLLM framework on an NVIDIA A100 (80GB) GPU to ensure stable, high-throughput, and efficient inference.
\FloatBarrier
\section{Prompts}
\label{app:prompts}
\subsection{Answer Generation}
\label{app:ag}
For generating answers, we provide five retrieved documents, user profile and question in the prompt.

% \begin{figure*}[!t]
\begin{tcolorbox}[colback=gray!10!white, colframe=black, title=Prompt for Answer Generation, width=\textwidth, enlarge left by=0mm, boxrule=0.8pt, colbacktitle=black!30!white]
You are tasked with writing a recommendation/fact-based answer that answers the user’s question based on a provided list of research abstracts and contextual information. Your response must:
\\

1. Directly address the user's concern, ensuring the answer is supported by the provided literature.

2. Incorporate the user's profile like timeline, professional background, Location, and concerns into the recommendations.

3. Clearly connect insights from the abstracts to the user's specific context and goals.

4. Make sure to output in points (1,2,3..) without inserting any **.

5. End your response with a confidence score (in percentage) and a short explanation for that score.
\\
Here are the 5 research abstracts:

1. \textcolor{blue}{\{lit1\}}  \\
2. \textcolor{blue}{\{lit2\}} \\ 
3. \textcolor{blue}{\{lit3\}} \\  
4. \textcolor{blue}{ \{lit4\}} \\  
5. \textcolor{blue}{\{lit5\}} \\

Context: \textcolor{blue}{\{context\}}  \\
Question: \textcolor{blue}{\{question\}} \\ 

Based on the above abstracts, write the answer in points. Make sure to take into account all the information in the context like profession, timeline, etc. Do not include subpoints.
\end{tcolorbox}
% \end{figure*}
% \twocolumn

\subsection{Specificity}
\label{app:sp}
For specificity evaluation, we provide claim, retrieved documents from the knowledge base and specific details to verify.
\begin{figure*}[!t]
\begin{tcolorbox}[colback=gray!10!white, colframe=black, title=Prompt for Specificity Evaluation, width=\textwidth, enlarge left by=0mm, boxrule=0.8pt, colbacktitle=black!30!white]
You are a strict evaluator of specificity and factuality.

Given for each claim: \\
                    - A factual claim\\
                    - A list of evidence passages from a trusted source \\
                    - A set of specific details extracted from the claim (hazard type, location, timeline, intensity) \\
Your task is to evaluate the claim using ONLY the provided evidence.\\
                    % ----------------------------\\
                    \textbf{LABEL DEFINITIONS}\\
                    % ----------------------------\\
                    For each specific detail (hazard, location, timeline, intensity), use EXACTLY one of the following labels:\\
                    - "yes":\\
                    The detail is explicitly mentioned in the claim AND
                    it matches the same specific detail discussed in the knowledge source.\\
                    - "no":\\
                    The detail is explicitly mentioned in the claim BUT
                    the knowledge source does NOT provide sufficient information to verify it.
                    (This includes cases where the evidence contradicts the claim or does not confirm it.)\\
                    - "N/A":\\
                    The detail is NOT mentioned in the claim at all.\\
                    For location matching, agreement at the STATE level is sufficient; an exact county or city match is not required.
                    Do NOT infer or assume any facts beyond the evidence.
                    Lack of verification MUST be labeled as "no" (not "N/A").\\
                    % ----------------------------\\
                    \textbf{EVALUATION STEPS}\\
                    % ----------------------------\\
                    Your task is to:\\
                    1. Determine whether the **claim is factually true, false, or partially correct**, using ONLY the evidence.\\
                    2. For each of the 4 specific details (hazard, location, timeline, intensity):\\
                     - Assign "yes", "no", or "N/A" based on the rules above.\\
                     - Provide a brief factual explanation for your decision.\\
                    3. Justify your overall factuality decision concisely and objectively.\\
                    4. If the claim is "true", cite the exact evidence passage(s) that support it.\\
                    5. If the claim is "false" or "partially correct", explain precisely which details are unsupported or incorrect.\\
                    % ----------------------------\\
                    \textbf{OUTPUT FORMAT}\\
                    % ----------------------------\\
                    Return your answer as a SINGLE JSON object in the following format (with no markdown, no extra text, and no explanations outside the JSON):\\ 
                    \{\{
                    "claim": "<Claim>",\\
                    "hazard": "yes" | "no" | "N/A",\\
                    "hazard\_reasoning": "<Explain whether hazard mentioned in the claim is explicitly supported>",\\
                    "location": "yes" | "no" | "N/A",\\
                    "location\_reasoning": "<Explain whether location mentioned in the claim is supported>",\\
                    "timeline": "yes" | "no" | "N/A",\\
                    "timeline\_reasoning": "<Explain whether timeline like date and range of years mentioned in the claim is supported>",\\
                    "intensity": "yes" | "no" | "N/A",\\
                    "intensity\_reasoning": "<Explain whether intensity mentioned in the claim is supported>"\\
                    \}\}\\
                    % ----------------------------\\
                    \textbf{INPUTS}\\
                    % ----------------------------\\
                    Claim: \textcolor{blue}{\{claim\}}  \\
                    Specific Details to Check:
                    \textcolor{blue}{\{specific\_info\}}  \\
                    Evidence Passages: \textcolor{blue}{\{knowledge\}}  \\

\end{tcolorbox}
\end{figure*}
\FloatBarrier
\subsection{Answer Relevance}
We use LLM for masking important specific details for answer relevance evaluation. 
\label{app:ar}

% \begin{figure*}[!t]
\begin{tcolorbox}[colback=gray!10!white, colframe=black, title=Prompt for Masking, width=\textwidth, enlarge left by=0mm, boxrule=0.8pt, colbacktitle=black!30!white]
You are a semantic masker. Given the following answer, replace: \\

- hazard types with [HAZARD] \\
- profession-related terms with [PROFESSION] \\
- concern with [CONCERN] (e.g., critical vulnerabilities, maintenance strategies, modernization measures, maintenance strategies, projected impact, design standards, cascading impacts etc.) \\
- infrastructure with [INFRASTRUCTURE] (e.g., "highway network", "bridge system", "public transit system", "railway infrastructure", "airport facilities", "port facilities", "freight terminals", "traffic control systems", "water treatment plant", "wastewater system", "dam infrastructure", "stormwater system", "coastal protection", "water distribution network", "electrical grid", "power distribution network", "EV charging network", "renewable energy infrastructure", "energy storage facilities", "power transmission lines", "substations", "public buildings", "critical facilities", "commercial structures", etc.) \\

Keep the structure natural and readable.

Answer: \textcolor{blue}{\{answer\}}
\\

\end{tcolorbox}
% \end{figure*}

% \section{Example Appendix}
\label{sec:appendix}

\end{document}